\documentclass[10pt,journal,compsoc]{IEEEtran}
\ifCLASSOPTIONcompsoc
  \usepackage[nocompress]{cite}
\else
  \usepackage{cite}
\fi

\ifCLASSINFOpdf
\else
\fi

\pdfoutput=1

\usepackage{graphicx}
\usepackage[utf8]{inputenc} 
\usepackage[T1]{fontenc}    
\usepackage{url}            
\usepackage{booktabs}       
\usepackage{amsfonts}       
\usepackage{nicefrac}       
\usepackage{microtype}      
\usepackage{epsfig}
\usepackage{float}
\usepackage{multicol}
\usepackage{multirow}
\usepackage{amssymb}
\usepackage{times}
\usepackage{amsmath}
\usepackage{wrapfig}
\usepackage{xspace}
\usepackage{color}
\usepackage{colortbl}
\usepackage[dvipsnames, svgnames, x11names]{xcolor}
\usepackage[misc]{ifsym}
\usepackage{makecell}
\usepackage{soul}
\usepackage{bm}
\usepackage{wrapfig}
\usepackage{graphicx}
\usepackage{bbding}
\usepackage{xcolor}
\usepackage{setspace}
\usepackage{subcaption}
\usepackage{threeparttable}

\definecolor{linkcolor}{RGB}{255,0,0}
\definecolor{urlcolor}{RGB}{255,105,180}
\definecolor{citecolor}{RGB}{66,168,235}
\usepackage[pagebackref,breaklinks=true,colorlinks,citecolor=blue,urlcolor=blue,linkcolor=blue,bookmarks=false]{hyperref}
\usepackage{adjustbox}
\usepackage{pifont}
\usepackage{caption}
\usepackage{array}
\usepackage{enumitem}
\newcolumntype{C}[1]{>{\centering\arraybackslash}p{#1}}

\definecolor{lightgray}{rgb}{0.8, 0.8, 0.8}
\definecolor{lgray}{rgb}{0.66, 0.66, 0.66}
\definecolor{whit_tab}{RGB}{255, 255, 255}
\definecolor{gray_tab}{RGB}{235, 235, 235}
\definecolor{oran_tab}{RGB}{254, 247, 241}
\definecolor{blue_tab}{RGB}{200, 227, 245}
\definecolor{lblu_tab}{RGB}{231, 239, 248}

\usepackage[ruled,vlined,linesnumbered]{algorithm2e}

\usepackage[capitalize]{cleveref}
\crefname{section}{Sec.}{Secs.}
\Crefname{section}{Section}{Sections}
\Crefname{table}{Table}{Tables}
\crefname{table}{Tab.}{Tabs.}

\newlength\savewidth

\renewcommand{\paragraph}[1]{\vspace{1.25mm}\noindent\textbf{#1}}

\newcommand{\modelname}{CE3D\textsuperscript{++}}

\newcommand{\revisioncolor}[1]{\textcolor{black}{#1}}

\begin{document}

\title{Chat-Edit-3D\textsuperscript{++}: Interactive 3D and 4D Scene Editing via Large Language Models}

\author{Shuangkang Fang, Yufeng Wang, Yi-Hsuan Tsai, Wenrui Ding, Yi Yang, Shuchang Zhou, Ming-Hsuan Yang
    
\IEEEcompsocitemizethanks{
\IEEEcompsocthanksitem Shuangkang Fang is with the School of Electrical and Information Engineering, Beihang University, Beijing 100191, China. Email: skfang@buaa.edu.cn.
\IEEEcompsocthanksitem Yufeng Wang and Wenrui Ding are with the Institute of Unmanned System, Beihang University, Beijing 100191, China. Email: \{wyfeng, ding\}@buaa.edu.cn.
\IEEEcompsocthanksitem Yi-Hsuan Tsai is with Atmanity Inc., USA. Email: wasidennis@gmail.com.
\IEEEcompsocthanksitem  Yi Yang and Shuchang Zhou are with Megvii Research, Beijing 100190, China. Email: \{yangyi, zsc\}@megvii.com.
\IEEEcompsocthanksitem Ming-Hsuan Yang is with the University of California at Merced, Merced. Email: mhyang@ucmerced.edu.
}
\thanks{}
\thanks{Corresponding authors: Yufeng Wang and Ming-Hsuan Yang}
}

\IEEEtitleabstractindextext{
\begin{abstract}
Recent work on image content manipulation based on vision-language pre-training models has been effectively extended to text-driven 3D scene editing.
However, existing schemes for 3D scene editing still have certain shortcomings, hindering their further development as interactive design tools.
Such schemes typically adhere to fixed input patterns, limiting flexibility in text input. 
Furthermore, their editing capabilities are constrained by a single or a few 2D visual models and require intricate pipeline design to integrate these models into 3D reconstruction processes.
To address the aforementioned issues, we propose the Hash-Atlas network, which reformulates 3D scene editing as operations on 2D atlas images, thereby achieving a workflow decoupling of the 2D editing and 3D reconstruction processes.
Building on this foundation, we introduce a dialogue-based 3D scene editing approach, termed \modelname{}, which is centered on a large language model (LLM) that allows arbitrary textual input from users and interprets their intentions, subsequently facilitating the autonomous invocation of the corresponding visual models.
Additionally, we extend \modelname{} to monocular 4D scenes by imposing motion constraints on moving objects and further fine-tuning the LLM by creating a trajectory dataset related to editing tasks, which enables the smaller LLM to schedule up to 30 different visual tools accurately.
Experimental results demonstrate that \modelname{} effectively integrates multiple visual models to achieve diverse visual editing effects, possessing strong scene comprehension and multi-round dialog capabilities.
The source codes and trained models are available at \url{https://github.com/Fangkang515/CE3D}. 

\end{abstract}
}

\maketitle

\IEEEdisplaynontitleabstractindextext

\IEEEpeerreviewmaketitle

\begin{figure*}[t]
    \centering
    \includegraphics[width=1.\textwidth]{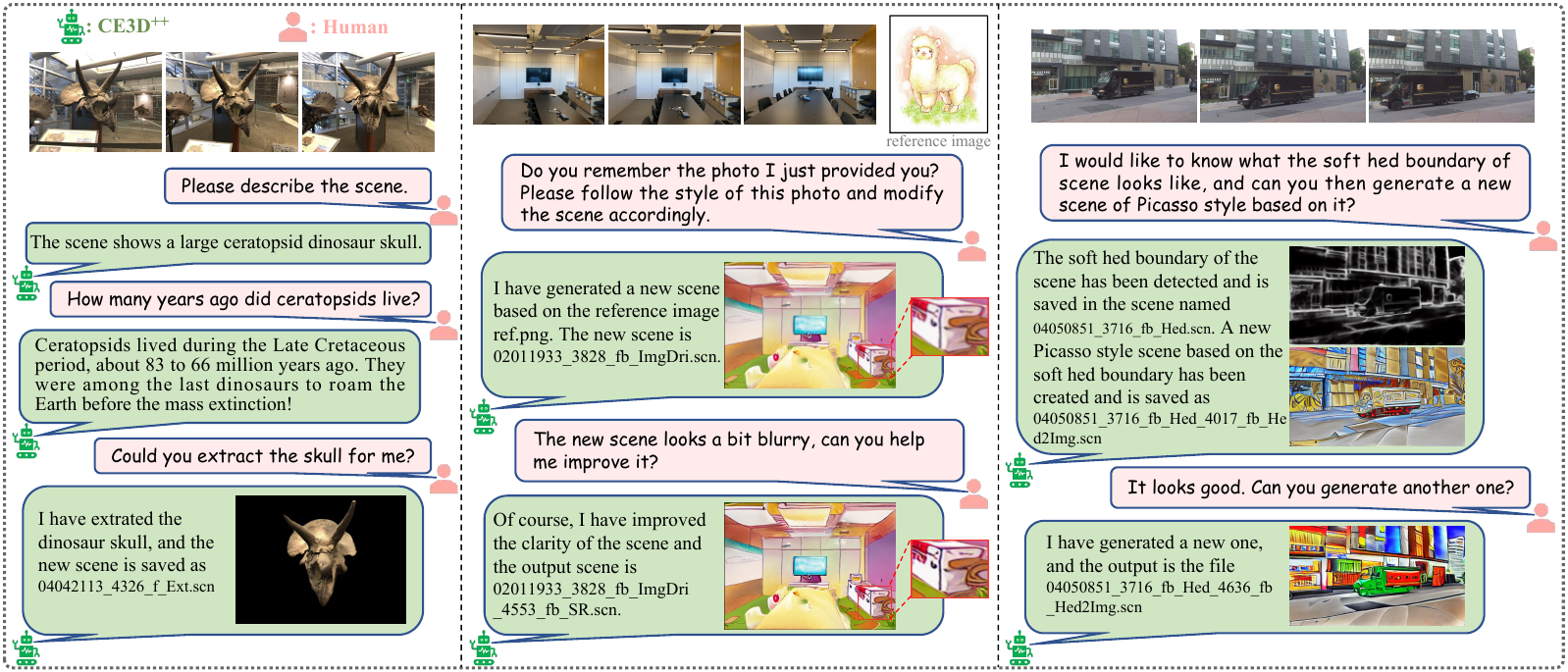}
    \caption{\textbf{Examples of chatting with \modelname{}}. We propose \modelname{}, a novel paradigm for 3D and 4D scene editing, which is compatible with a variety of extant visual models. By managing these visual tools through the LLMs, we achieve challenging dialogue-based scene editing tasks that are difficult to accomplish with previous methods.}
    \label{fig:shocking}
\end{figure*}
\begin{figure}[ht]
    \centering
    \includegraphics[width=0.48\textwidth]{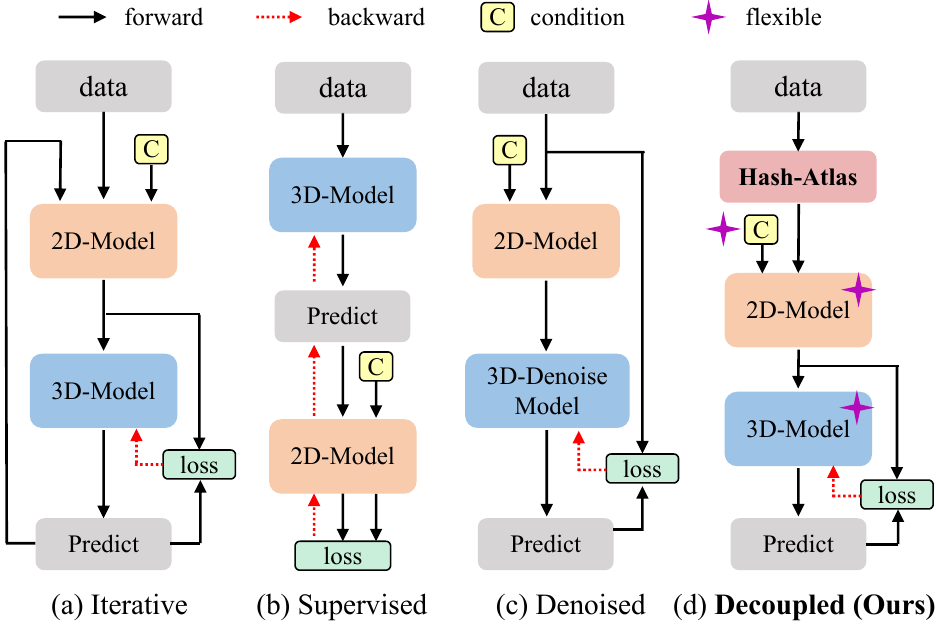}
    \caption{\textbf{Differences between \modelname{} and other typical text-driven editing frameworks}. Existing methods (a)-(c) require complex integration of specific 2D and 3D models. While \modelname{} (d) decouples the 2D editing process from the 3D representation, enhancing its compatibility with various visual models.}
    \label{fig:diff_pipe}
\end{figure}

\section{Introduction} \label{sec:intro} 
The rise of advanced dialogue systems based on large language models (LLMs)~\cite{brown2020gpt3, Driess2023_PaLM,touvron2023_llama,wei2022chain-1}, such as ChatGPT~\cite{chatGPT3.5,achiam2023gpt-4} and Gemini~\cite{Gemini1.0,Gemini1.5}, marks a new era in human-computer interaction. 
These systems, trained with massive datasets, can comprehend complex contexts, engage in logical reasoning, and generate coherent responses.
They have been used to address practical issues while significantly reducing human effort.
In addition to text processing capabilities, recent research has successfully applied LLMs to the 2D visual domain.
Examples like Visual-ChatGPT~\cite{wu2023visual-gpt} and MM-REACT~\cite{yang2023mm-react} leverage LLMs as logical hubs, facilitating the integration and management of multiple visual models, including detection, segmentation, and visual question answering (VQA), allowing more natural interactions with users.

Motivated by the success of LLMs, this work exploits LLMs for 3D and 4D scene editing tasks (see~\cref{fig:shocking}).
Nevertheless, constructing such a dialogue system is challenging mainly because open-ended dialogue implies diverse user queries and demands, which a single or a few editing approaches cannot adequately address. 
One potential solution is to integrate multiple visual editing methods.
However, existing text-driven 3D editing frameworks, such as those based on NeRF~\cite{mildenhall2020nerf,fang2025nerf-gs,haque2023instruct,fang2023DN2N,wang2022clipnerf,wang2022nerfart,yin2023or-nerf} or Gaussian-splatting~\cite{kerbl20233dgaussian,fang2026dropan-sh,fang2023gaussianeditor,ye2023gaussian-seg-edit}, constrain this development. 
\cref{fig:diff_pipe} shows the typical frameworks:
(a) Iterative methods, such as IN2N~\cite{haque2023instruct} and Gaussian-Editor~\cite{fang2023gaussianeditor} require iterative embedding of information from 2D visual models into the edited 3D scenes. (b) Supervised approaches~\cite{yin2023or-nerf,wang2022clipnerf} rely on 2D models to supervise the 3D reconstruction process. 
(c) Denoised methods, e.g., DN2N~\cite{fang2023DN2N}, necessitate the prior modeling of 3D inconsistency between different views caused by the 2D editing model. 
These coupled designs, shown in (a)-(c), are typically intricate and require identifying well-suited 2D and 3D models that function effectively within their frameworks. 
This makes it difficult to flexibly integrate diverse visual models and hinders the deployment of LLMs for 3D image editing.

In this work, we propose a novel editing paradigm based on workflow decoupling, which separates the 2D editing stage from the 3D reconstruction pipeline, as shown in~\cref{fig:diff_pipe}\textcolor{blue}{(d)}. 
\revisioncolor{This paradigm is not tied to predefined visual models and is broadly compatible with diverse 2D or 3D visual models.}
To this end, a neural network is initially trained to map the different views of a 3D scene to plane atlases, thereby transforming 3D editing into 2D atlas-space operations.
\revisioncolor{The atlas representation naturally accommodates a wide range of 2D visual models}, eliminating the need for additional designs to adapt 2D models to the 3D reconstruction process. 
This ensures that our approach remains highly compatible and flexible when integrating various visual tools. 
Building upon this foundation, we develop a dialogue system named \modelname{}, which enables LLMs to accurately parse user text inputs and manage multiple visual models and associated scene files in a text manner, achieving the execution of complex 3D scene editing tasks through an interactive dialogue interface.

We further extend our method to monocular 4D scene editing by selecting the pivot frame with the largest visible range of the moving target to apply positional constraints within the atlas and introduce a motion loss to mitigate image quality degradation caused by motion occlusion. 
To handle complex user instructions and manage a large array of visual tools, we construct an LLM trajectory-tuning dataset tailored to editing tasks in this study.
This dataset fills a gap in the field, enabling lightweight LLMs to support more than 30 different vision tools and accurately interpret intricate user commands.

The main contributions of our paper are:
\begin{itemize}
    \item We develop the Hash-Atlas method, which translates the editing of 3D and 4D scenes into manipulation of 2D atlases, thereby decoupling the editing and reconstruction processes to avoid the intricacies of conventional pipeline architectures.
    \item Using LLMs, we propose a dialogue framework for editing 3D and 4D scenes, termed \modelname{}, which encompasses a mechanism for parsing user input text and formulating user responses, an executor for implementing atlas editing logic, and managing multiple visual models and scene files. In addition, we construct a trajectory-tuning dataset to enhance the reliability of LLMs.
    \item Extensive experimental results demonstrate that \modelname{} exhibits strong scalability and is compatible with various existing 2D, 3D and 4D visual models. Compared to previous methods, \modelname{} facilitates more robust text parsing, richer editing, and more natural interactions.
\end{itemize}

The preliminary results of this work are presented in a conference paper~\cite{ce3d}. 
This work differs in four aspects. 
(1) The previous work demonstrated feasibility for static 3D scenes, but was not well compatible with dynamic 4D scenes. 
To address this, we introduce a pivot motion loss into the Hash-Atlas model, which improves the reliability of dynamic object tracking and enables the high-quality mapping of monocular 4D scenes to 2D atlases.
(2) We expand the number of supported visual models from 20 to 30+, significantly enriching the available editing functionalities.
(3) We develop a trajectory-tuning dataset to improve smaller LLMs to parse complex user instructions and execute multi-step editing tasks, thereby enhancing their practical deployment value.
(4) We conduct extensive experiments, including comparisons with 4D scene editing methods, demonstrations of enhanced editing functionalities, and an analysis of the reliability of different LLMs in the editing process.

\section{Related Work} \label{sec:related_works}
\noindent{\textbf{Vision-Language Pretrained Models.}}
Significant progress in large-scale vision-language pre-training~\cite{radford2021clip,li2022blip,li2023blip2,dhariwal2021diffusion,rombach2022high,nichol2021glide,kawar2022imagic,ho2020denoising,saharia2022photorealistic,ho2022classifier} has enabled models to effectively understand complex visual and textual data. 
These pre-trained models have been widely used to improve the performance of downstream tasks. 
For example, combining CLIP with diffusion~\cite{dhariwal2021diffusion,ho2020denoising,song2020denoising,saharia2022photorealistic} allows high-fidelity image generation based on text~\cite{rombach2022high,ho2022classifier}.
More relevant to this work are methods for controllable image generation and editing.
For example, ControlNet~\cite{zhang2023controlnet} generates new images based on conditions like text, depth, and edge.
Instruct-Pix2Pix~\cite{brooks2022instructpix2pix} facilitates instruction-based image editing, and BLIP-diffusion~\cite{li2023blip-diffusion} enables style transfer based on reference images.

\vspace{1mm}
\noindent{\textbf{LLMs for Vision Tasks.}} 
LLMs~\cite{brown2020gpt3,chatGPT3.5,achiam2023gpt-4, Driess2023_PaLM,touvron2023_llama,claude3_report,team2023gemini,deepseek-v3,qwen} significantly demonstrate the advanced capabilities of novel language models, serving as a basis for research and development of dialogue systems.
However, LLMs lack the inherent ability to process visual input directly. 
One solution involves mapping visual information into text~\cite{hu2022-l2v-1, wang2022-l2v-2, yang2022empirical-l2v-3, zeng2022socratic-l2v-4,fang2025meshllm}, which enables subsequent processing by LLMs.
Another solution~\cite{wu2023visual-gpt,yang2023mm-react,suris2023vipergpt} leverages the chain-of-thought capability~\cite{wei2022chain-1,brown2020gpt3}, treating LLMs as an inference hub and calling on relevant visual tools based on the user's text input to handle various visual tasks. 
Several methods have been developed to apply LLMs to 3D scenes, such as Chat-3D~\cite{wang2023chat3D} and 3D-GPT~\cite{sun20233d-gpt}. 
Chat-3D focuses primarily on interpreting scene content and addressing tasks such as VQA and image captioning, while 3D-GPT emphasizes using LLMs for the generation of 3D scenes. 
In this paper, we are mainly concerned with exploring how LLMs can assist in editing 3D and 4D scenes.

\vspace{1mm}
\noindent{\textbf{3D Scene Editing.}}
3D Scene Editing involves both the scene reconstruction and editing processes. 
Earlier methods are based on structural representations such as point clouds and meshes~\cite{huang2021learning,mu20223d,zhou2014color,hollein2022stylemesh,han2021exemplar,cai2026geometry,SFF}. 
Recently, scene editing techniques based on NeRF~\cite{mildenhall2020nerf} have been developed.
Examples include leveraging the implicit modeling nature of NeRF for editing~\cite{martin2021nerfinthewild, kobayashi2022decomposing,tang2022CCNeRF,li2022climatenerf, uav-enerf}, employing structural transformations to achieve diverse editing capabilities~\cite{fang2022pvd,fang2023pvdal}, and utilizing images as references for style transfer~\cite{huang2022stylizednerf,gu2021stylenerf,zhang2022arf,boss2021nerd,boss2021neural,mu20223dphotostylization}.
Integration with pre-trained vision language models has yielded promising results in 3D scene editing research, as shown in CLIP-NeRF~\cite{wang2022clipnerf}, IN2N~\cite{haque2023instruct}, Nerf-Art~\cite{wang2022nerfart}, and DN2N~\cite{fang2023DN2N}, among others. 
Beyond NeRF, Gaussian-splatting~\cite{kerbl20233dgaussian} has also also been applied to 3D scene editing~\cite{fang2023gaussianeditor,ye2023gaussian-seg-edit}.
However, these 3D scene editing methods based on NeRF or Gaussian-splatting heavily depend on intricate pipelines combining 3D reconstruction with 2D pre-trained models, complicating component modification and replacement. 
Additionally, their rigidity in input modes hinders their ability to adapt to different text expressions.

\vspace{1mm}
\noindent{\textbf{4D Scene Editing.}}
Unlike 3D scenes, representing 4D scenes requires capturing the dynamic structures of objects. 
Some methods extend NeRF or Gaussian splatting by incorporating deformation modeling, enabling an effective representation of dynamic scenes~\cite{dynamic-nvs,monocular-dynamic-nvs,neural3d-video,nerfies,park2021hypernerf,dynamic-3dgs,deformable-3dgs,real-4dgs}. 
Building on these approaches and existing 3D scene editing methods, recent efforts have been made in 4D scene editing.  
For instance, 4D-Editor~\cite{4D-Editor} achieves localized scene editing by combining semantic distillation with user-provided strokes. 
Control4D~\cite{control4d} employs Gaussian Tri-planes to represent 4D scenes and utilizes GANs~\cite{GAN} for scene manipulation. 
Instruct-4D-to-4D (IN4Dto4D)~\cite{instruct4Dto4D} enhances frame-wise consistency in edited scenes by introducing an anchor-aware attention module, while CTRL-D~\cite{CTRL-D} propagates single-frame edit effects across an entire scene by fine-tuning Instruct-Pix2Pix~\cite{brooks2022instructpix2pix}. 
Although these methods demonstrate notable progress, they remain constrained by predefined 4D scene representations and 2D editing models, lacking the flexibility to adapt freely or interpret diverse text prompts from users.

\begin{figure*}[tp]
    \centering
    \begin{minipage}{1.\textwidth}
        \centering
        \begin{figure}[H]
            \centering
            \includegraphics[width=1.\textwidth]{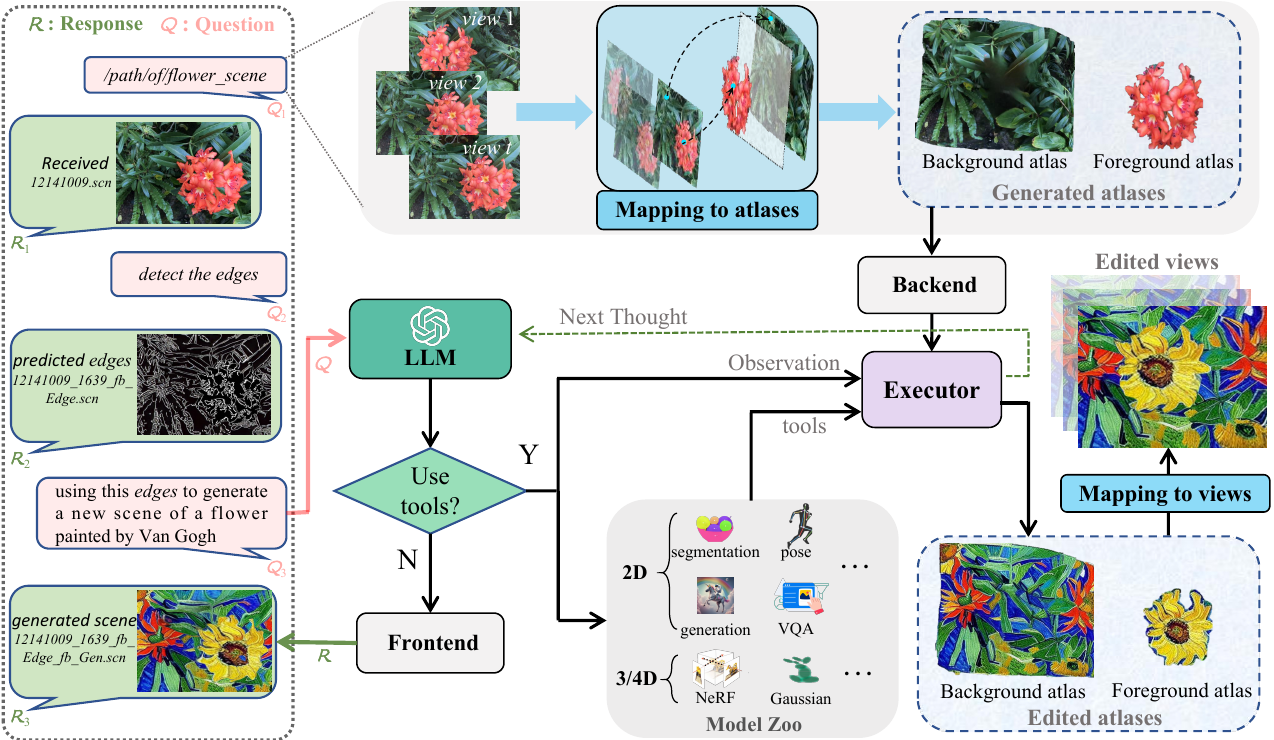}
            \caption{\textbf{Illustration of the \modelname{} framework}. The basic process is as follows: (1) Given the user's text query, LLM interprets the text and determines whether visual tools are required for this dialogue.
            (2) When visual tools are needed, LLM will call the desired tools from the model zoo and provide them with the corresponding parameters. 
            (3) The Backend further queries the atlases and other files to be invoked. In addition, if the atlases do not exist, the Backend first acquires them using the Hash-Atlas network.
            (4) The Executor executes visual tools to edit the atlases and feeds back new status to LLM for subsequent actions. The edited atlases are then mapped back to the 3D or 4D scene views through the Hash-Atlas network for later scene reconstruction.
            (5) Since one dialogue may require multiple model calls, LLM repeats the above process until it determines that visual tools are no longer needed. Then the Frontend responds to the user with the editing results and LLM's outputs.
            }
            \label{fig:overview_pipe}
        \end{figure}
        \vspace{-3mm} 
        \begin{figure}[H]
            \centering
            \includegraphics[width=1.\textwidth]{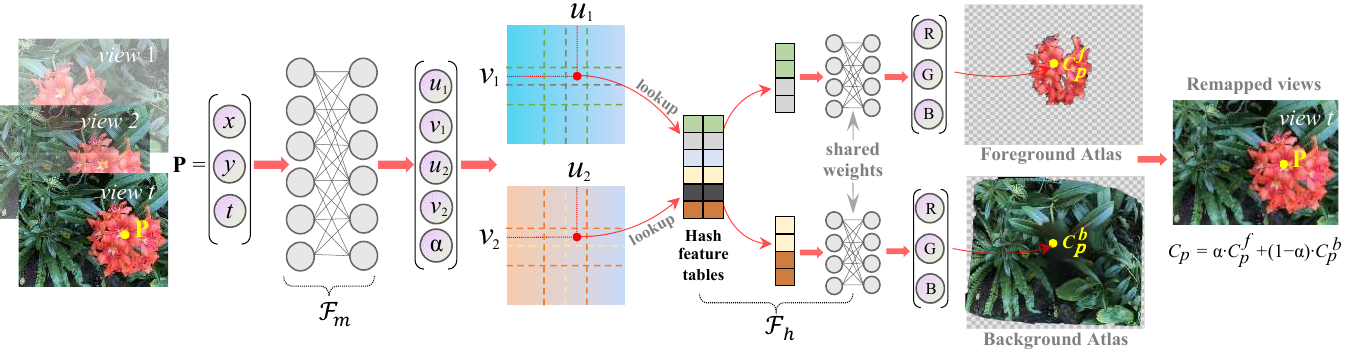}
            \caption{\textbf{Illustration of Hash-Atlas}. Given multi-views of a 3D or 4D scene, $F_m$ first maps each pixel coordinate from the views to two UV spaces and predicts the transparency of the foreground atlas. Then, $F_h$ predicts RGB values at each coordinate in the UV space, yielding foreground and background atlases. When the atlases are edited, they can be mapped back to the original scene views.
            }
            \label{fig:hash_atlas}
        \end{figure}
    \end{minipage}
\end{figure*}

\section{Proposed Method} \label{sec:method}
We first illustrate the \modelname{} overview pipeline (\cref{fig:overview_pipe}), followed by the design of the Hash-Atlas network (Section~\ref{subsec:hash-atlas}), editing strategy in atlas space (Section~\ref{subsec:editing_atlas}), and components of the dialogue system within \modelname{} (Section~\ref{subsec:dialog_system}).

\subsection{Hash-Atlas Model} \label{subsec:hash-atlas}
We present an approach that directly maps images from various views of a scene onto 2D foreground and background atlases (as shown in~\cref{fig:hash_atlas}), relocating the 3D and 4D scene editing process to be executed in 2D space. %
Similar techniques are initially used to map video frames to atlases~\cite{kasten2021layered-atlas1, geyer2023tokenflow-atlas2, bar2022text2live-atlas3}, which require continuous frames and smooth camera motion, unlike the 3D scene data used in this paper. 
To achieve the desired editing capabilities outlined in this article, the atlases should meet the following criteria:
(1) Distortion and skew in the atlas should be minimized to maintain visual content. (2) Foreground and background atlases should be approximately aligned for accurate editing.
(3) A faster and more precise mapping is required to facilitate efficient editing.

\vspace{1mm}
\noindent{\textbf{Hash-Atlas Formulation.}}
To meet the aforementioned conditions, we devise a network based on a hash structure~\cite{muller2022instant}, as illustrated in~\cref{fig:hash_atlas}.
Assuming there are $T$ views in the scene, the point $P=[x, y, t]$ at the $t$-th view is mapped to two different UV coordinates using the function $F_m$:
\begin{equation}
    [u_1, v_1, u_2, v_2, \alpha] = F_m(x, y, t),
\label{eq:p2uv}
\end{equation}
where $(u_1, v_1)$ and $(u_2, v_2)$ represent the coordinates in the two UV spaces. 
The parameter $\alpha$, which ranges from 0 to 1, signifies the weight of the pixel value in the foreground atlas.
Then $F_h$ is used to predict the RGB values corresponding to the foreground and background atlases in the UV coordinate:
\begin{equation}
    C_p^f = F_h(u_1, v_1); \quad C_p^b = F_h(u_2, v_2),
\label{eq:uv2atlas}
\end{equation}
where $F_h$ incorporates a hash structure~\cite{muller2022instant} to capture texture details in images and enable faster model training and inference. 
To share the weights of $F_h$ for two different UV coordinates, $(u_1, v_1)$ is arranged to the interval $[0, 0.5]$ and $(u_2, v_2)$ to the interval $[0.5, 1]$.
After obtaining the pixel values $C_p^f$ and $C_p^b$ in atlases, the original pixel value in the scene's view for point $P$ can be reconstructed:
\begin{equation}
    C_p = \alpha \cdot C_p^f + (1-\alpha) \cdot C_p^b.
\label{eq:rec_p}
\end{equation}
When atlases are edited, each view of the scene with editing effects can be restored by~\cref{eq:rec_p} without retraining the Hash-Atlas network again. 

\begin{figure*}[tp]
    \centering
    \includegraphics[width=1.\textwidth]{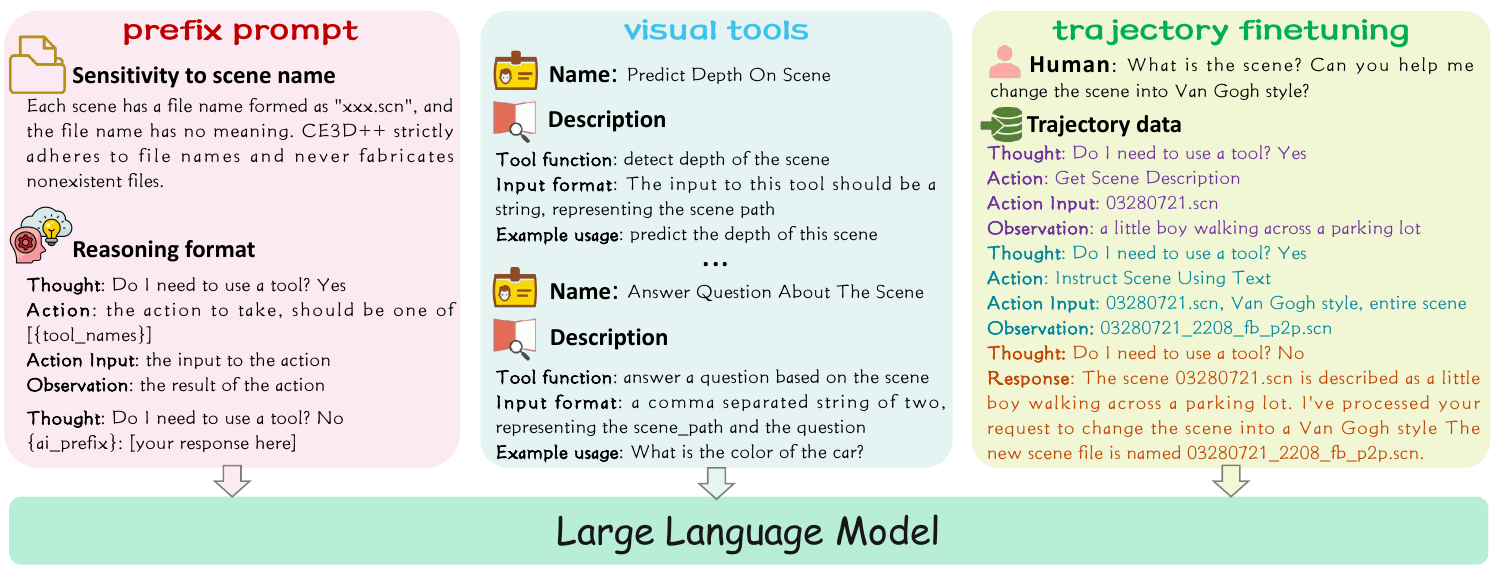}
    \caption{\textbf{Illustration of the customized LLM agent}. We show example prompts for configuring an LLM as an agent, including the linguistic management of multiple non-linguistic file formats within a scene, the specification of thought processes, and the integration of external visual tool invocation knowledge. 
    Additionally, we construct a trajectory-tuning dataset for tool invocation, which is utilized to fine-tune the LLM, thereby enhancing its comprehension of complex instructions and improving its reliability. We simplify some text to fit into the page. Please refer to our code for the complete prompts.}
    \label{fig:trajectory}
    \vspace{-0mm}
\end{figure*}

\vspace{1mm}
\noindent{\textbf{Training and Loss Terms.}}
To ensure that the obtained atlases appear more natural, avoiding excessive tilting and distortion of objects, during the early stages of model training, we exclusively use $P=[x,y,0]$ from the $0$-th view. 
Then, the pre-training positional loss is defined as follows:
\begin{equation}
    \mathcal{L}_{pos} = ||x-u_1|| + ||x-u_2|| + ||y-v_1|| + ||y-v_2||.
\end{equation}
This loss encourages minimal changes in the position of the scene from the $0$-th view after the coordinate mapping.

Furthermore, pre-training for \(\alpha\) involves initially determining what the foreground of the scene is by a VQA model~\cite{li2022blip} and its corresponding mask by a
segmentation model~\cite{kirillov2023sam, liu2023grounding-dino}. 
Assuming the foreground mask for point \(P\) is \(m_p\), the loss for pre-training \(\alpha\) is defined as follows:
\begin{equation}
    \mathcal{L}_{\alpha} = \text{CE}(\alpha_p, m_p) + ||(1-\alpha_p) C_{p}^{f}||,
\end{equation}
where \(\text{CE}\) represents the cross-entropy loss. 
The second term encourages the \(\alpha\) and the foreground atlas to be sparse, which facilitates a distinct separation between the contents of the foreground and background atlases.

After the pre-training phase, the entire model can be trained by supervising the reconstruction view from the atlases.
Nevertheless, we note that this training induces significant omissions in the background atlas, adversely affecting later editing processes.
To address this issue, we introduce the inpainting loss. 
Initially, we use the ProPainter model~\cite{zhou2023propainter} to perform preliminary inpainting on the masked background, generating a new set of inpainted views. 
Assuming that the point \(P\) in the original view corresponds to the \(\tilde{P}\) in the inpainted view, the reconstruction loss can be expressed as follows:
\begin{equation}
    \mathcal{L}_{rec} = \mathcal{L}_{rec}^{ori} + \mathcal{L}_{rec}^{pro} = || C_p - GT(C_p) || + ||C_{\tilde{p}}^b - GT(C_{\tilde{p}}^b) ||,
\end{equation}
where \(GT(\cdot )\) denotes the ground truth obtained from the original or inpainted views of the scene. 
In addition, similar to~\cite{kasten2021layered-atlas1}, we incorporate rigid and flow constraints in the scene. 
The purpose of $\mathcal{L}_{rigid}$ is to maintain the relative spatial location between different points without drastic changes. 
At the same time, the $\mathcal{L}_{flow}$ encourages mapping the corresponding points from different views to the same location on the atlas. Therefore, the total loss can be expressed as follows:
\begin{equation}
    \mathcal{L}_{total} = \mathcal{L}_{init} + \mathcal{L}_{rec}^{ori} + \mathcal{L}_{rec}^{pro} + \mathcal{L}_{rigid} + \mathcal{L}_{flow},
    \label{eq:3D_total_loss}
\end{equation}
where $\mathcal{L}_{init} = \mathcal{L}_{pos} + \mathcal{L}_{\alpha}
$ is only employed during the initial training phase.

\begin{figure*}[tp]
    \centering
    \begin{minipage}{\textwidth}
        \centering
        \begin{table}[H]
            \centering
            \caption{\textbf{Quantitative comparisons with LNA~\cite{kasten2021layered-atlas1} for atlas}. We report metrics that are averaged over various scenes from different datasets, achieving a 1.1$\sim$5.1dB improvement in PSNR, a 14.2$\sim$18.6x acceleration in training, and a 7.6$\sim$9.0x increase in FPS. Our method provides higher-quality atlases for the subsequent editing process and ensures faster mapping from atlases to scene views.}
            \resizebox{0.95\textwidth}{!}{\begin{tabular}{c|cc|cc|cc|cc|cc|cc}
\toprule
      & \multicolumn{2}{c|}{\textbf{LLFF}} & \multicolumn{2}{c|}{\textbf{TanksAndTemples (3D)}} & \multicolumn{2}{c|}{\textbf{IBRNet-collect (3D)}} & \multicolumn{2}{c|}{\textbf{CE3D-collect (3D)}} & \multicolumn{2}{c|}{\textbf{Dycheck (4D)}} & \multicolumn{2}{c}{\textbf{DynamicNeRF (4D)}} \\
\cmidrule{2-13}      & LNA   & Hash-Atlas & LNA   & Hash-Atlas & LNA   & Hash-Atlas & LNA   & Hash-Atlas & LNA   & Hash-Atlas & LNA   & Hash-Atlas \\
\midrule
PSNR$\uparrow$ & 23.04  & \textbf{24.66} & 19.24 & \textbf{20.4} & 25.56 & \textbf{28.53} & 23.96 & \textbf{27.24} & 22.67 & \textbf{27.73} & 24.32 & \textbf{28.26} \\
SSIM$\uparrow$ & 0.67  & \textbf{0.79 } & 0.57  & \textbf{0.61} & 0.77  & \textbf{0.87} & 0.63  & \textbf{0.84} & 0.61  & \textbf{0.85} & 0.74  & \textbf{0.84} \\
LPIPS$\downarrow $ & 0.40  & \textbf{0.17 } & 0.59  & \textbf{0.46} & 0.31  & \textbf{0.11} & 0.53  & \textbf{0.12} & 0.58  & \textbf{0.14} & 0.35  & \textbf{0.14} \\
Train Time (h)$\downarrow $ & 10.75  & \textbf{0.60 } & 14.69 & \textbf{1.03} & 9.75  & \textbf{0.56} & 14.02 & \textbf{0.75} & 14.16 & \textbf{0.82} & 11.52 & \textbf{0.69} \\
Inference FPS$\uparrow $ & 0.66  & \textbf{5.93} & 0.25  & \textbf{2.11} & 0.79  & \textbf{6.13} & 0.67  & \textbf{5.57} & 0.72  & \textbf{6.17} & 0.71  & \textbf{5.94} \\
\bottomrule
\end{tabular}%

}
            \label{tab:compare_atlas}
        \end{table}
        \vspace{-1mm}
        \begin{figure}[H]
            \centering
            \includegraphics[width=1.\textwidth]{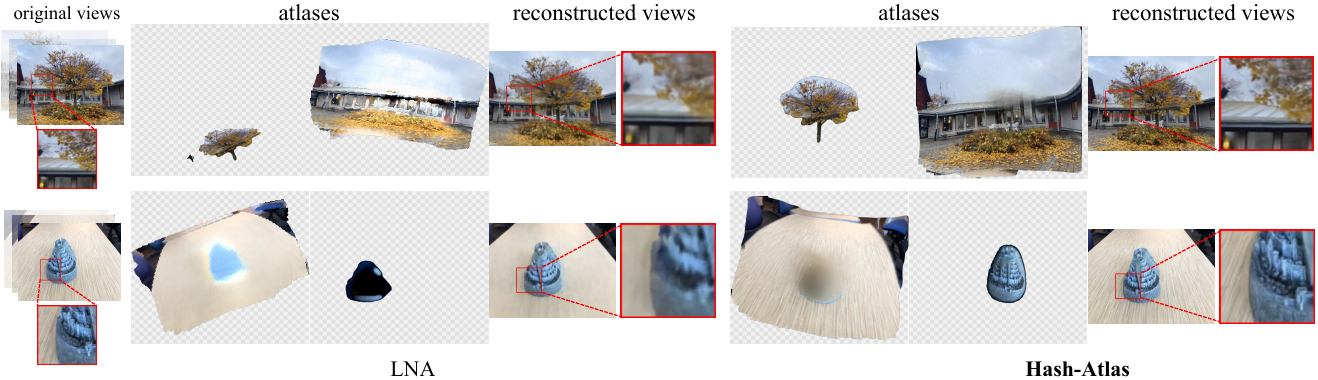}
            \caption{\textbf{Qualitative comparisons with LNA~\cite{kasten2021layered-atlas1} for atlas}. The proposed Hash-Atlas method retains more scene details, keeps the relative position of the objects in foreground and background atlases largely unchanged, and minimizes distortion. This provision of more precise scene target locations and more commonsensical visual information to subsequent editing models enables better editing outcomes.}
            \label{fig:compare-atlas}
        \end{figure}
    \end{minipage}
\end{figure*}

\begin{figure}[th]
    \centering
    \includegraphics[width=0.48\textwidth]{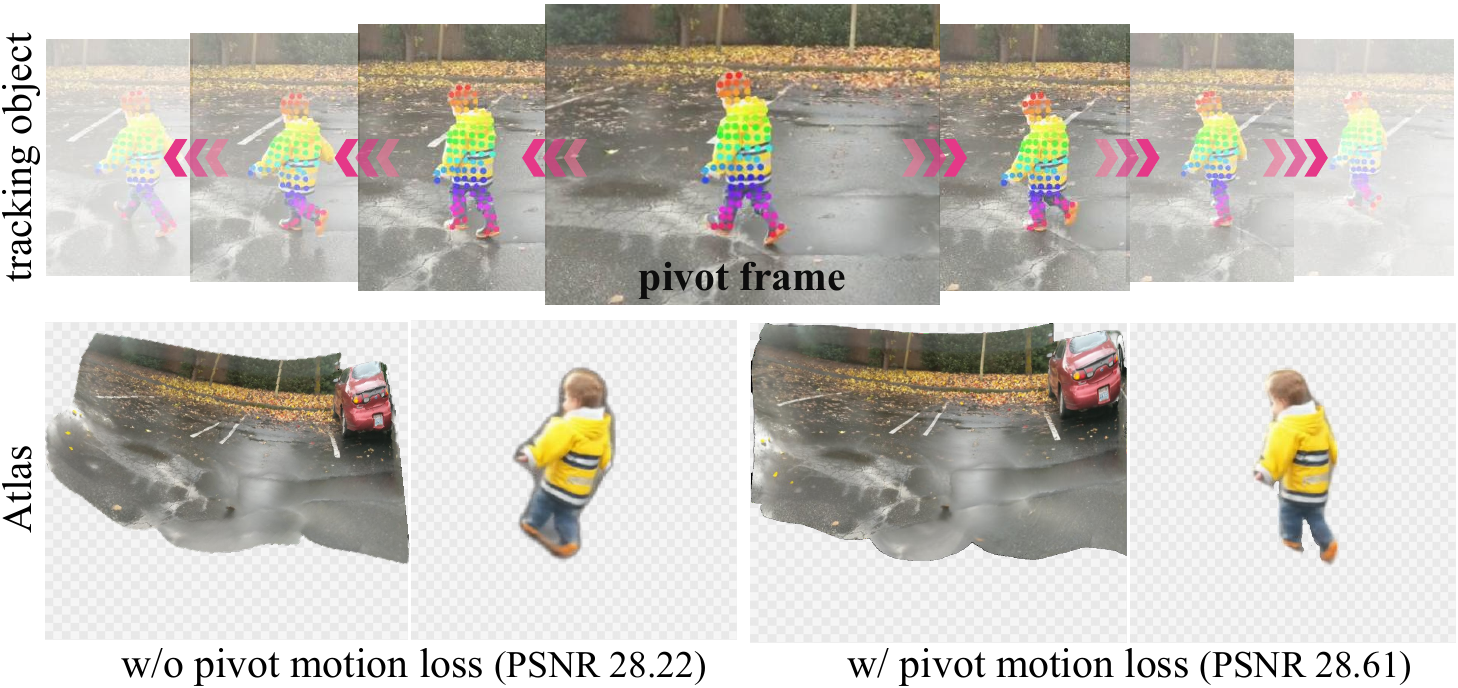}
    \caption{\textbf{Effectiveness of pivot-motion loss}. For comparison, we enlarge the object in the foreground atlas.}
    \label{fig:4d-motion-loss}
\end{figure}

\begin{figure*}[!ht]
    \centering
\includegraphics[width=0.95\textwidth]{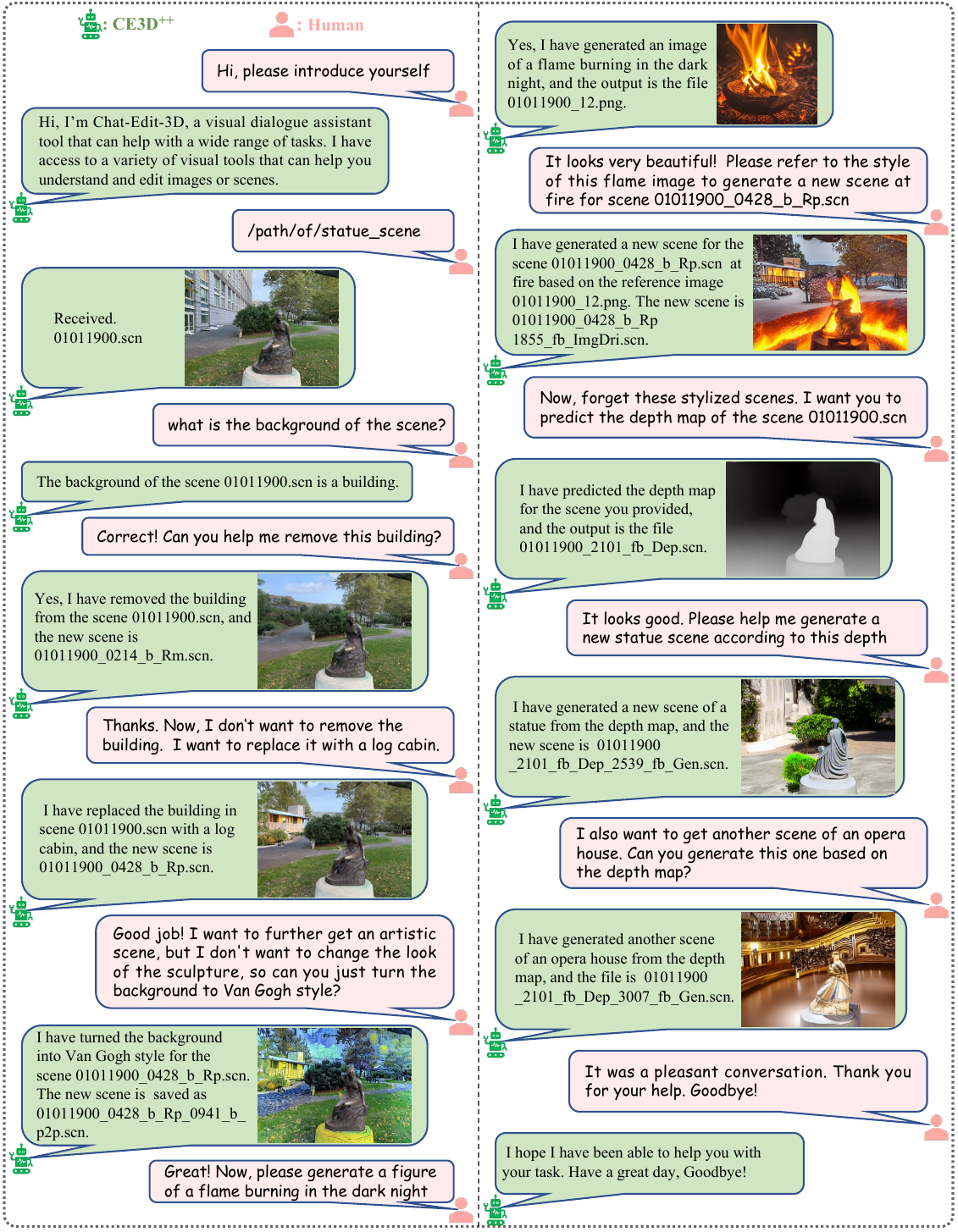}
    \caption{\textbf{Multi-round dialogue editing results}. \modelname{} performs editing of 3D scenes on the 2D atlases, \revisioncolor{thus facilitating broad compatibility with a wide range of existing 2D models and supporting diverse editing capabilities}. Moreover, by utilizing LLMs for the management of various visual models and parsing user text inputs, \modelname{} enables users to engage in more flexible text inputs and easily achieve multi-round dialogue.
    }
    \label{fig:chat-1}
\end{figure*}

\begin{figure*}[tp]
  \centering
  \includegraphics[width=0.7\textwidth]{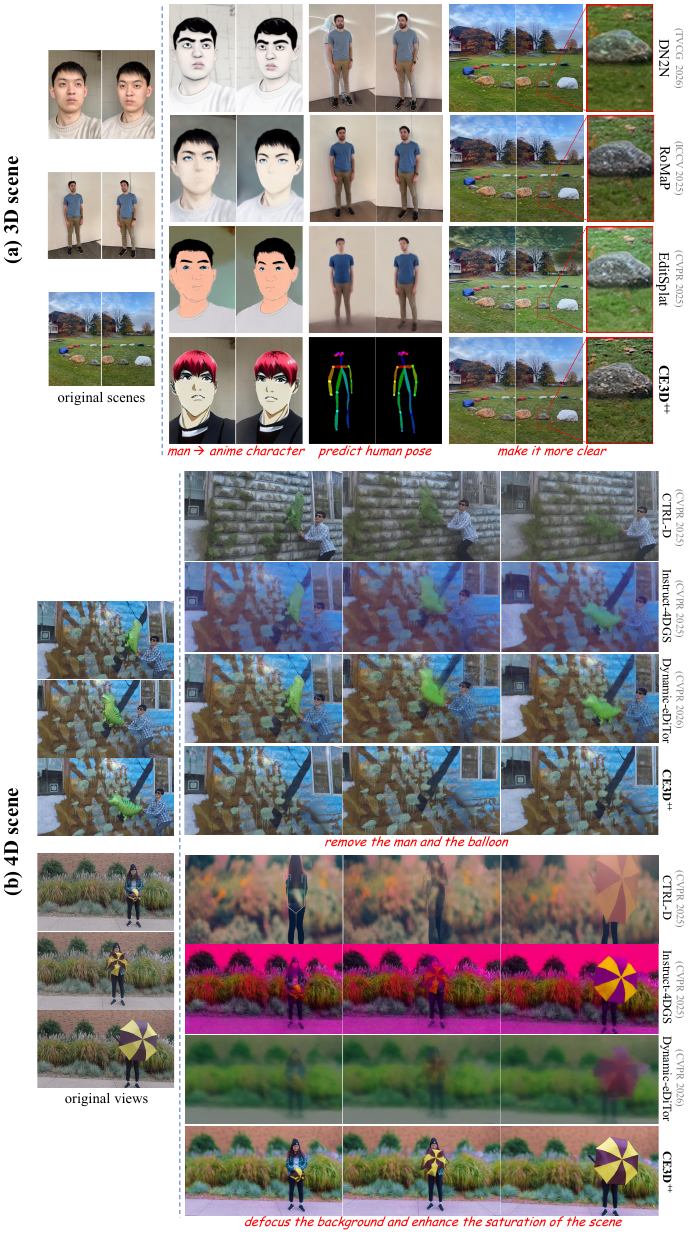}
\caption{\textbf{Comparison with other 3D and 4D scene editing methods}. Existing state-of-the-art methods exhibit limited editing capability. In contrast, the proposed \modelname{} provides \revisioncolor{broad compatibility with a wide range of 2D visual models, enabling the integration of diverse editing functionalities}.}
\label{fig:compare_editing}
\end{figure*}

\begin{figure*}[t]
    \centering
    \subfloat[\textbf{Comparisons for text-query diversity.} We present qualitative results for three text queries of the same editing type. Although all three editing instances involve edge detection, IN2N produces notably varied results when the accompanying text varies. While \modelname{} possesses a robust language comprehension capability that mitigates instability in the editing outcomes resulting from textual discrepancies.
    ]{
        \includegraphics[width=0.47\textwidth]{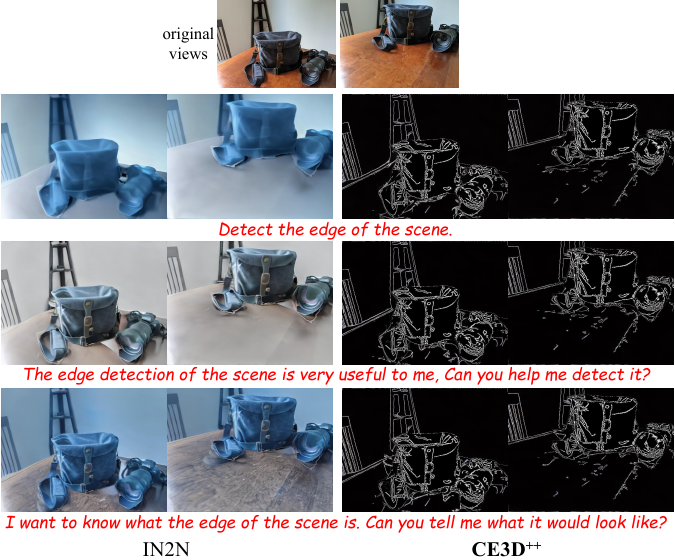} 
        \label{fig:compare_text_multi}
    }
    \hfill
    \subfloat[\textbf{Comparisons with IN2N for multi-round dialogue}. IN2N lacks sufficient editing capabilities and falls short in text parsing, making it challenging to handle complex multi-round dialogue tasks. By introducing a decoupled novel editing framework, our method effectively addresses these challenges and outperforms IN2N significantly.
    ]{
        \includegraphics[width=0.48\textwidth]{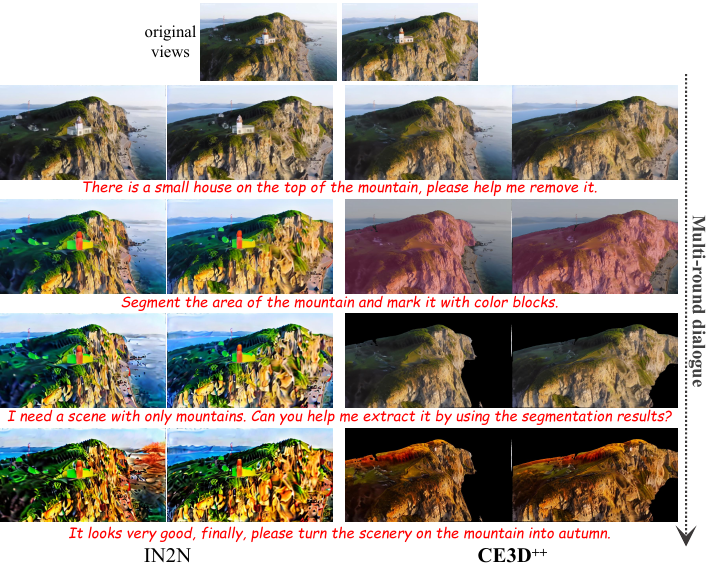} 
        \label{fig:compare-multi-round}
    }
    \caption{\textbf{Comparison with IN2N~\cite{haque2023instruct} for text-query diversity and multi-round dialogue}.}
    \label{fig:compare-IN2N}
\end{figure*}

\vspace{1mm}
\noindent{\textbf{Pivot Motion Loss for 4D Scene.}}
In contrast to 3D scenes, employing the discussed method on 4D dynamic scenes causes significant distortion and information loss in the derived 2D atlases due to object movement (illustrated in~\cref{fig:4d-motion-loss}), thereby impairing editing efficacy.
To mitigate this degradation, an additional loss constraint is imposed on the atlas mapping of moving objects.  
First, to ensure that the mapped foreground atlas retains sufficient structural and texture information of the dynamic object, we analyze the object masks in different views and select the view with the highest pixel count as the Pivot-View to initialize the position of the object in the atlas, denoted as \( \mathcal{L}_{\text{init}}^{\text{pivot}} \).  
Then, we perform uniform sampling on the moving object within the Pivot-View to obtain \(N\) key points \( \{ p_i \}_{i=1}^N \).
For each point \( p_i \), we utilize CoTracker3~\cite{cotracker3} to track its corresponding positions across adjacent frames before and after Pivot-View in the temporal sequence. Let \( p_i^t \) denote the tracked position of \( p_i \) in frame \( t \). 
%
To ensure temporal consistency, we design a motion loss that enforces different tracked points \( p_i^t \) to map to identical 2D atlas coordinates. 
Specifically, for successfully tracked points $p_i^t$, we define the motion loss as follows: 
\begin{equation}
    \mathcal{L}_{\text{motion}} = \sum_{i=1}^N \sum_{\substack{t \in \mathcal{T}_i}} \left\| F_m(p_i) - F_m(p_i^{t}) \right\|,
    \label{eq:4D_motion_loss}
\end{equation}
where \( \mathcal{T}_i \) represents the set of frames where key point \( p_i \) is successfully tracked. This loss helps reduce tracking inconsistencies and ensures the stability of moving object representations in the 2D atlas space.

\subsection{Editing in Atlas Space}\label{subsec:editing_atlas}
We observe that directly editing two atlases independently and then mapping them back to the scene views usually does not yield satisfactory editing results. 
This is mainly because a single atlas contains incomplete scene information, especially in a sparse foreground atlas. 
This restriction hinders visual models from fully understanding scene semantics, thus compromising editing accuracy.
Therefore, we design a merge-split strategy for editing atlases.
In this process, we leverage the parsing of LLMs and a VQA model~\cite{li2022blip} to identify editing areas. 
If the regions involve only the background, editing is performed directly on the background atlas. 
Otherwise, we overlay the foreground atlas on the background, serving it as the actual atlas for editing. 
Subsequently, we use the original foreground mask and the new object mask to separate the edited atlas.
We use ``Executor'' to represent the actual editing process, as shown in Fig.~\ref{fig:overview_pipe}.

\subsection{Dialog System} 
\label{subsec:dialog_system}
\noindent{\textbf{Sensitivity to Scene Names.}}
As a language model, LLM cannot directly access information outside of text.
Considering the numerous files in the editing workflow, inputting all as text into the LLM is impractical.
Consequently, we represent the involved files with a single string formatted as ``xxx.scn.'' 
This string is unique and meaningless to prevent LLM from fabricating scene names, as illustrated in~\cref{fig:trajectory}.
Although this scene name is not a genuinely readable file, further processing by the Frontend and Backend, as shown in~\cref{fig:overview_pipe}, allows \modelname{} to handle real files effectively. 
The Frontend assembles user replies from both edited results and LLM outputs, whereas the Backend orchestrates the allocation of scene files during editing and oversees the management of file names for new scenes.

\vspace{1mm}
\noindent{\textbf{Reasoning for User Queries.}} When presented with user input, the LLM simulates a thought process: ``Do I need to use a visual tool?'' to ``Which tools do I need?'' to ``What should be the specific input for the tools?''
Therefore, it is essential to pre-inject information about each visual tool into LLMs to complete this reasoning process. 
Similar to \cite{wu2023visual-gpt,yang2023mm-react}, we annotate each visual tool with four categories as depicted in~\cref{fig:trajectory}: the tool name, under what circumstances it should be used, the required parameters, and specific input examples.

\vspace{1mm}
\noindent{\textbf{Trajectory-tuning for LLMs.}}
The above design aligns the generalization capabilities of LLMs with domain-specific visual tools, enabling automated execution of complex 3D and 4D scene editing tasks. 
In our experiments, typical LLMs, such as ChatGPT, are capable of completing these tasks effectively. However, lightweight and cost-effective LLMs may face challenges. 
As the number of visual tools increases, the pre-injected domain knowledge required for LLMs grows significantly. The increasingly complex tool descriptions gradually diminish the accuracy of tool invocation. 
Furthermore, in multi-step editing tasks, LLMs may make erroneous decisions during intermediate steps, leading to the final task failure. To address these issues, we develop a trajectory-tuning dataset tailored to our editing task with only 1,000 dialogue samples. 
This data set significantly improves the accuracy of LLMs in orchestrating external tools.
As illustrated in~\cref{fig:trajectory}, our dataset design focuses mainly on visual tool attributes, reasoning behind scheduling decisions, and decision-making processes. 
This targeted approach improves the LLMs' understanding of proper tool usage. 
In particular, the fine-tuned model preserves its scheduling accuracy when new visual tools are introduced, even when these tools were not included in the original training data.

\section{Experiments and Analysis} 
\label{sec:exp}
We first quantitatively (Table~\ref{tab:compare_atlas}) and qualitatively (Fig.~\ref{fig:compare-atlas}) show the performance of our Hash-Atlas model against the LNA scheme~\cite{kasten2021layered-atlas1}.
Subsequently, we demonstrate the diverse editing results of the proposed \modelname{} (Figs.~\ref{fig:shocking} and~\ref{fig:chat-1}), and comparisons with other 3D and 4D scene editing methods (Figs.~\ref{fig:compare_editing} and~\ref{fig:compare-IN2N}, and Table~\ref{tab:compare_clip}).
We then analyze the performance of various LLMs in our tasks and demonstrate the effectiveness of the trajectory-tuning strategy (Figs.~\ref{fig:compare_llm_rate} and~\ref{fig:compare_qwen}).
Finally, we present extensive ablation studies to validate our design choices (Fig.~\ref{fig:ablation} and Tables~\ref{tab:ablation} and~\ref{tab:rebuttal_4d_atlas_hyperParams} ) and limitations (Fig.~\ref{fig:failure_cases}) of our method. 

\subsection{Implementation Details}
\noindent\textbf{Datasets}. The datasets used in 3D scene editing include LLFF~\cite{mildenhall2019llff}, NeRF-Art~\cite{wang2022nerfart}, IN2N-collect~\cite{haque2023instruct}, IBRNet-collect~\cite{wang2021ibrnet}, TanksAndTemple~\cite{knapitsch2017tanks}.
We additionally collect a forward-facing outdoor dataset, denoted CE3D-collect, to supplement scene types. This dataset comprises 11 scenes, each containing 50 to 80 images. We use COLMAP~\cite{schoenberger2016colmap1} to estimate the camera poses for each scene.
For 4D scene editing, we use the monocular DyCheck~\cite{gao2022dycheck} and DynamicNeRF~\cite{gao2021dynamicNeRF} datasets for performance evaluation. 

\vspace{1mm}
\noindent\textbf{Hash-Atlas Structure and Training Details}.
We implement the Hash-Atlas model in PyTorch, with the optimizer as AdamW~\cite{adamw}. 
We construct $F_m$ as an eight-layer MLP with a width of 256, along with ReLU activations. 
For $F_h$, we set the 
base resolution, number of levels, per-level scale, hash table size, and feature dimensions to 
16, 16, 1.5, $2^{15}$, and 2 respectively.  
The MLP, followed by the hash tables, has two hidden layers with a width of 64. 
The activation function is ReLU for each hidden layer and tanh for the output layer.
The initial learning rates for $F_m$ and $F_h$ are 1e-3 and 1e-2, respectively. We run 100,000 steps with a batch size of 10,000 pixel points.
The $\mathcal{L}_{pos}$ loss is utilized solely for the initial 1,000 iterations, after which the $\mathcal{L}_{\alpha}$ loss is trained only for the first 30,000 iterations. 
The resolution of the generated atlases stands at $1000\times 1000$ pixels. 
More details can be found in the released source code.

\vspace{1mm}
\noindent\textbf{Dialogue System Settings}.
The 3D and 4D models available for access by LLMs are TensoRF~\cite{chen2022tensorf}, 3D Gaussian-splatting~\cite{kerbl20233dgaussian} and 4D Gaussian-splatting~\cite{4dgs}. 
The 2D visual tools including VQA~\cite{li2022blip},
Image Generation~\cite{rombach2022high},
Image Caption~\cite{li2022blip},
Super-Resolution~\cite{rombach2022high}
Text-driven Stylize~\cite{brooks2022instructpix2pix},
Image-driven Stylize~\cite{li2023blip2},
Segmentation~\cite{kirillov2023sam},
Low-Light Image Enhancement~\cite{retinexformer-lowlight},
Image Recoloring~\cite{huggingface-recolor},
Image Dereflection~\cite{hu2025dereflection},
Artistic Typography Generation~\cite{artistic-typography},
Defocus and Blur Control~\cite{bokehme-defocus},
Saturation Enhancement~\cite{opencv_library},
Ghibli Style Control~\cite{easycontrol_ghibli}.
We also use GPT-4o~\cite{openai_4o_image_generation} and Doubao~\cite{doubao_model} for editing experiments. 
In addition, we experiment with the models that transfer the image to Line~\cite{gu2022-img2line}, Edge~\cite{xu2017img2edge}, Hed~\cite{xie2015-img2hed-img2sketch}, Depth~\cite{ranftl2021-img2depth-normal-2}, Normal Map~\cite{ranftl2021-img2depth-normal-2}, Sketch~\cite{xie2015-img2hed-img2sketch}, Pose\cite{cao2017img2pose}, and reverse conversion using ControlNet~\cite{zhang2023controlnet}.

We use ChatGPT~\cite{achiam2023gpt-4} as the default LLM. 
To comprehensively evaluate the impact of different LLMs on dialogue system stability, we conduct extensive experiments with other LLMs, including commercial API-based models such as Claude~\cite{claude3_report}, Gemini~\cite{team2023gemini}, Doubao~\cite{doubao_model} and Moonshot~\cite{kimi_website}, as well as open-source models such as DeepSeek-v3~\cite{deepseek-v3}, LLaMA~\cite{touvron2023_llama} and Qwen~\cite{qwen}. 
We divide the constructed trajectory-tuning dataset into training and testing sets with a 9:1 ratio for fine-tuning and evaluating the LLaMA-8B and Qwen-14B models. 
In particular, the test set includes eight visual tools not present in the training set.
Full-parameter fine-tuning for LLMs is carried out on 32 A800 80GB GPUs for 4 epochs using the AdamW optimizer with a learning rate of 2e-5. While editing experiments are performed on 2 A800 80GB GPUs. 

\begin{figure}[ht]
    \centering
    \includegraphics[width=0.5\textwidth]{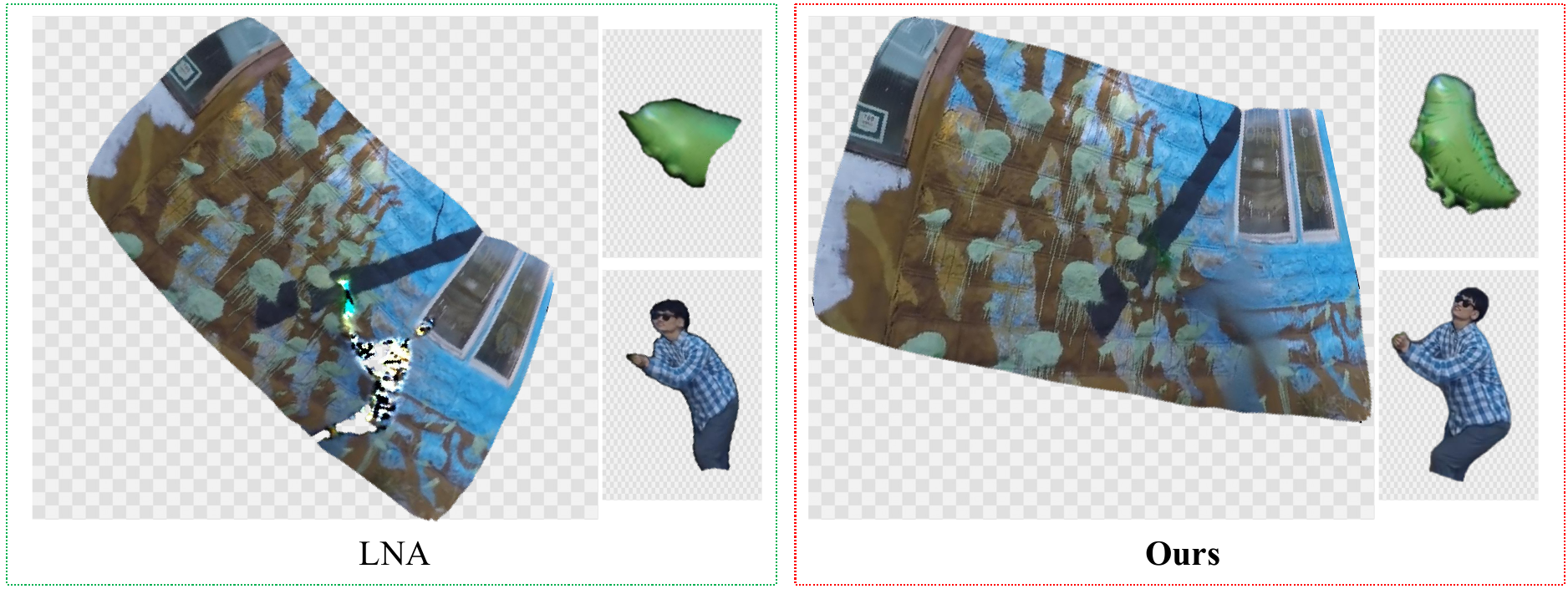}
\caption{\revisioncolor{{\textbf{Visualization of multi-layer atlas}. We extend Hash-Atlas to a three-layer setting. Compared with LNA, our method produces atlas representations with fewer visual artifacts.}}}
\label{fig:multi-layers}
\end{figure}

\begin{table}[ht]
    \centering
    \setlength{\tabcolsep}{1pt}
\caption{\revisioncolor{\textbf{Quantitative comparisons of editing results.}}}
    \resizebox{0.5\textwidth}{!}{
    \begin{tabular}{c|ccc|c|cc}
\toprule
\multirow{2}[4]{*}{\textbf{3D scene}} & LLFF  & CE3D-collect & NeRF-Art & \multirow{2}[4]{*}{\textbf{4D scene}} & DyCheck & DynamicNeRF \\
\cmidrule{2-4}\cmidrule{6-7}      & \multicolumn{3}{c|}{CLIP Similarity $\uparrow$} &       & \multicolumn{2}{c}{CLIP Similarity $\uparrow$} \\
\midrule
EditSplat~\cite{lee2025editsplat} & 0.188 & 0.212 & 0.201 & CTRL-D~\cite{CTRL-D} & 0.186  & 0.194 \\
RoMaP~\cite{kim2025RoMaP} & 0.224 & 0.240 & 0.229 & Instruct-4DGS~\cite{kwon2025Instruct-4DGS} & 0.223 & 0.251 \\
DN2N~\cite{fang2023DN2N} & 0.193 & 0.233 & 0.218 & Dynamic-eDiTor~\cite{lee2025dynamiceditor} & 0.236 & 0.267 \\
\textbf{\modelname{}} & \textbf{0.304} & \textbf{0.352} & \textbf{0.321} & \textbf{\modelname{}} & \textbf{0.313} & \textbf{0.320 } \\
\midrule
      & \multicolumn{3}{c|}{CLIP Directional Score $\uparrow$} &       & \multicolumn{2}{c}{CLIP Directional Score $\uparrow$} \\
\midrule
EditSplat~\cite{lee2025editsplat} & 0.121 & 0.153 & 0.136 & CTRL-D~\cite{CTRL-D} & 0.135  & 0.141 \\
RoMaP~\cite{kim2025RoMaP} & 0.145 & 0.181 & 0.159 & Instruct-4DGS~\cite{kwon2025Instruct-4DGS} & 0.157 & 0.158 \\
DN2N~\cite{fang2023DN2N} & 0.129 & 0.174 & 0.157 & Dynamic-eDiTor~\cite{lee2025dynamiceditor} & 0.141 & 0.174 \\
\textbf{\modelname{}} & \textbf{0.192} & \textbf{0.248} & \textbf{0.211} & \textbf{\modelname{}} & \textbf{0.193} & \textbf{0.214} \\
\midrule
      & \multicolumn{3}{c|}{Editing time (minutes) $\downarrow$} &       & \multicolumn{2}{c}{Editing time (minutes) $\downarrow$} \\
\midrule
EditSplat~\cite{lee2025editsplat} & 16.4  & 18.7  & 17.7  & CTRL-D~\cite{CTRL-D} & 66.8  & 35.4 \\
RoMaP~\cite{kim2025RoMaP} & 21.5  & 24.9  & 24.2  & Instruct-4DGS~\cite{kwon2025Instruct-4DGS} & 43.5  & 22.6 \\
DN2N~\cite{fang2023DN2N} & 26.5  & 36.3  & 31.5  & Dynamic-eDiTor~\cite{lee2025dynamiceditor} & 51.2  & 25.8 \\
\textbf{\modelname{}} & \textbf{5.6} & \textbf{8.8} & \textbf{8.4} & \textbf{\modelname{}} & \textbf{15.9} & \textbf{8.2} \\
\midrule
      & \multicolumn{3}{c|}{VRAM Peak (GB) $\downarrow$} &       & \multicolumn{2}{c}{VRAM Peak (GB) $\downarrow$} \\
\midrule
EditSplat~\cite{lee2025editsplat} & 14.7  & 14.9  & 14.5  & CTRL-D~\cite{CTRL-D} & 22.8  & 20.1 \\
RoMaP~\cite{kim2025RoMaP} & 16.3  & 17.2  & 16.6  & Instruct-4DGS~\cite{kwon2025Instruct-4DGS} & 24.6  & 22.7 \\
DN2N~\cite{fang2023DN2N} & \textbf{8.4} & \textbf{9.1} & \textbf{8.7} & Dynamic-eDiTor~\cite{lee2025dynamiceditor} & \textbf{16.8} & 15.5 \\
\textbf{\modelname{}} & 12.6  & 12.7  & 11.9  & \textbf{\modelname{}} & 17.4  & \textbf{14.2} \\
\bottomrule
\end{tabular}%

    }
    \label{tab:compare_clip}
\end{table}

\subsection{Atlas Reconstruction Results}
A high-quality atlas representation is critical to effective scene editing. 
Given the scarcity of existing methods for mapping 3D and 4D scene views onto atlases, we benchmark the proposed Hash-Atlas approach against LNA~\cite{kasten2021layered-atlas1}, a technique for constructing atlases in video.
In contrast to LNA, which relies on pure MLPs and assumes smooth inter-frame transitions, our method incorporates multi-resolution hash encoding, significantly accelerating both training and inference.
Furthermore, while LNA is prone to distortions under sparsely sampled and non-continuous viewpoint settings common in 3D scenes, our approach introduces dedicated pre-training and inpainting losses ($\mathcal{L}_{init}, \mathcal{L}_{rec}^{pro}$) to improve geometric stability and address occlusion-induced artifacts.
As presented in Table~\ref{tab:compare_atlas} and Fig.~\ref{fig:compare-atlas}, Hash-Atlas significantly outperforms LNA and exhibits superior adaptability for complex LLM-driven 3D and 4D editing operations.

\revisioncolor{The proposed Hash-Atlas representation naturally supports multi-layer dynamic scenes by dynamically adding atlas branches conditioned on object masks, with each branch corresponding to an independent object. 
As shown in Fig.~\ref{fig:multi-layers}, in a three-atlas setting, our method produces higher-quality atlases than LNA.}

\begin{figure}[t]
    \centering
    \includegraphics[width=0.485\textwidth]{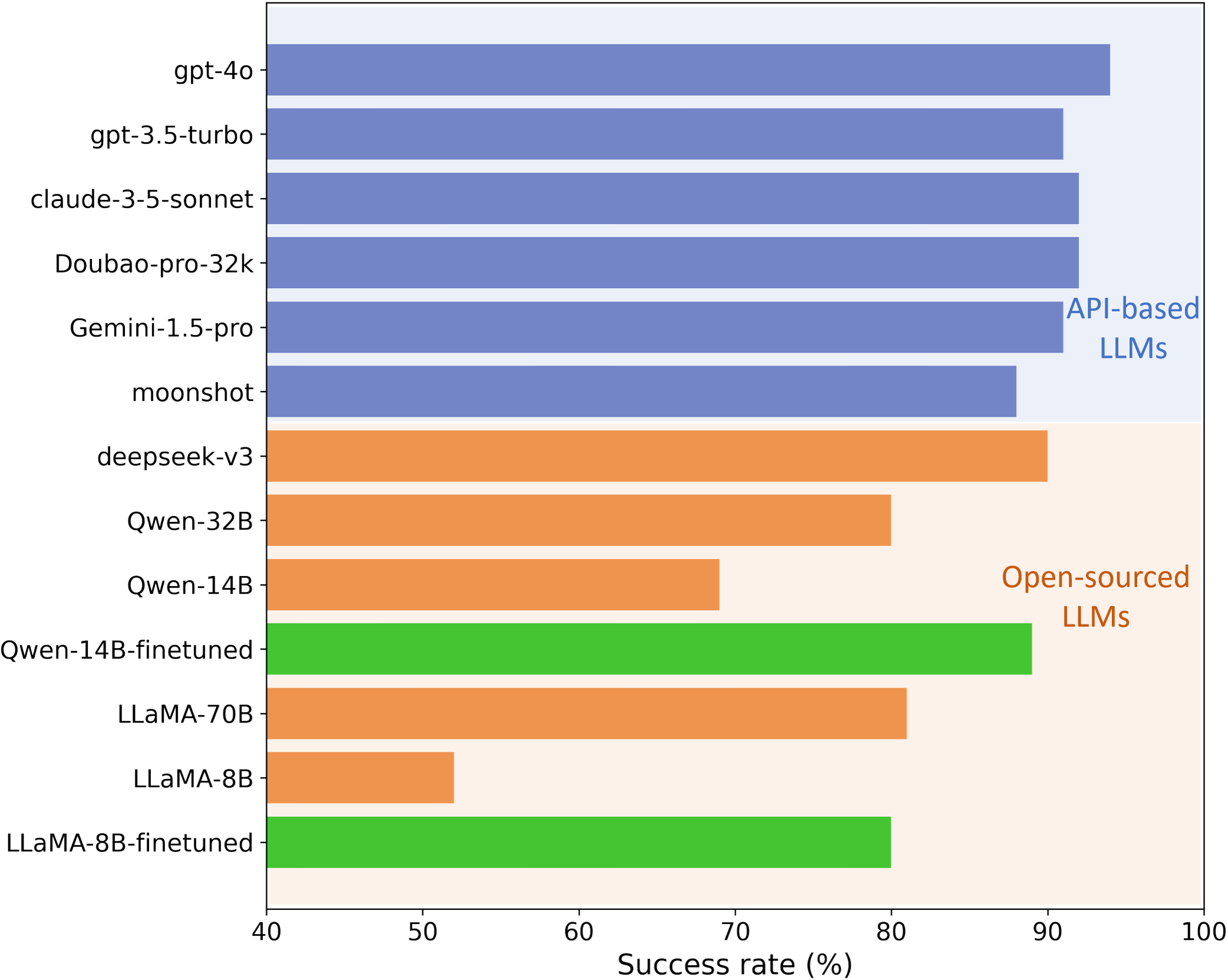}
    \caption{\textbf{Success rate of editing tasks for different large language models}. We evaluate the success rate of LLMs in accurately invoking external visual tools across 100 diverse complex user instructions. By fine-tuning Qwen-14B and LLaMA-8B with our curated trajectory data (shown in \textcolor[RGB]{69, 197, 50}{green}), we observe a significant improvement in success rates, demonstrating the effectiveness of this approach in enhancing the robustness of LLM-based agents.}
    \label{fig:compare_llm_rate}
\end{figure}

\begin{figure*}[t]
    \centering
    \includegraphics[width=1.0\textwidth]{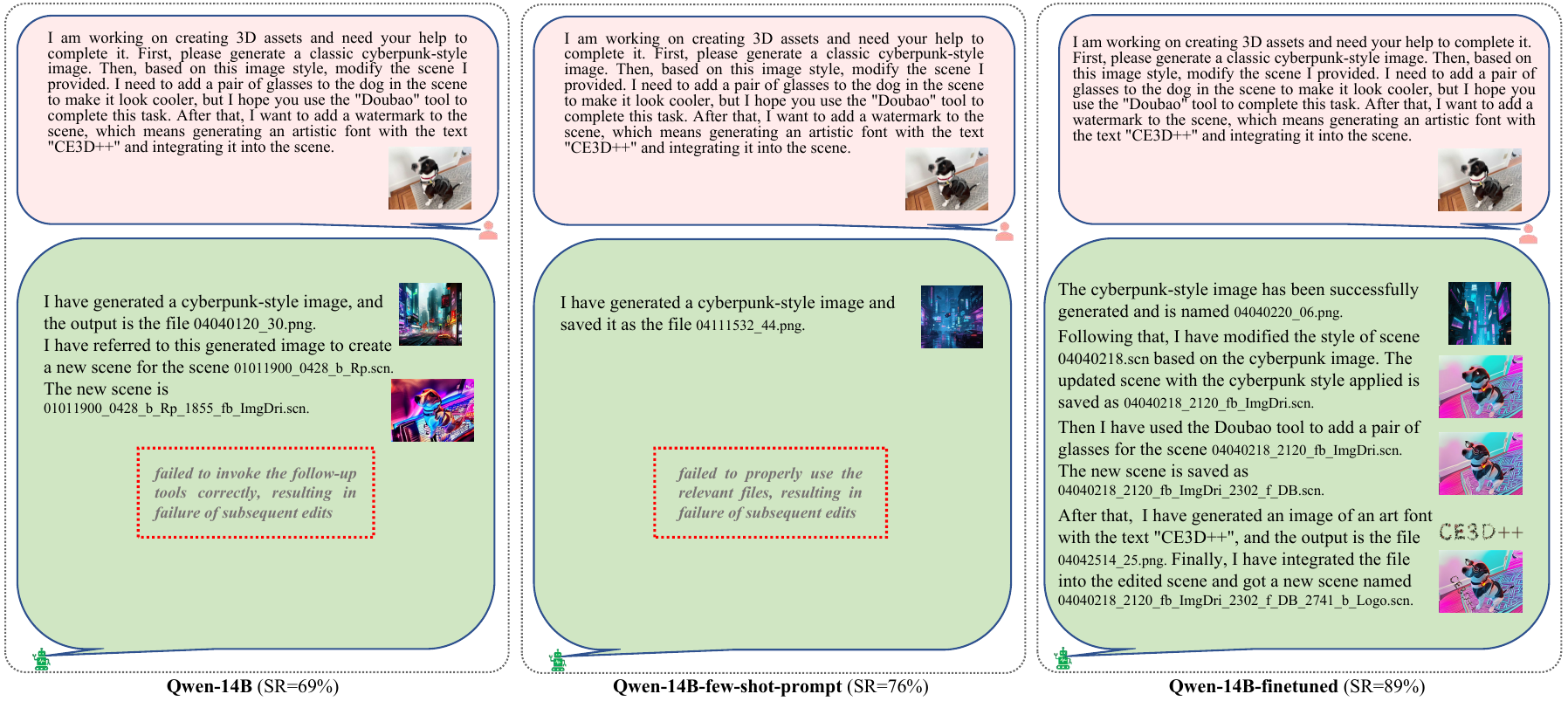}
\caption{\textbf{Comparison of LLM performance.} `SR' denotes the success rate under multiple editing instructions. Qwen-14B-finetuned
successfully invokes external tools to complete complex multi-step editing tasks.}
\label{fig:compare_qwen}
\end{figure*}

\begin{figure*}[t]
    \centering
    \subfloat[\textbf{Loss ablation on the Hash-Atlas network.} The text query is ``remove sofa''.]{
        \includegraphics[width=0.95\textwidth]{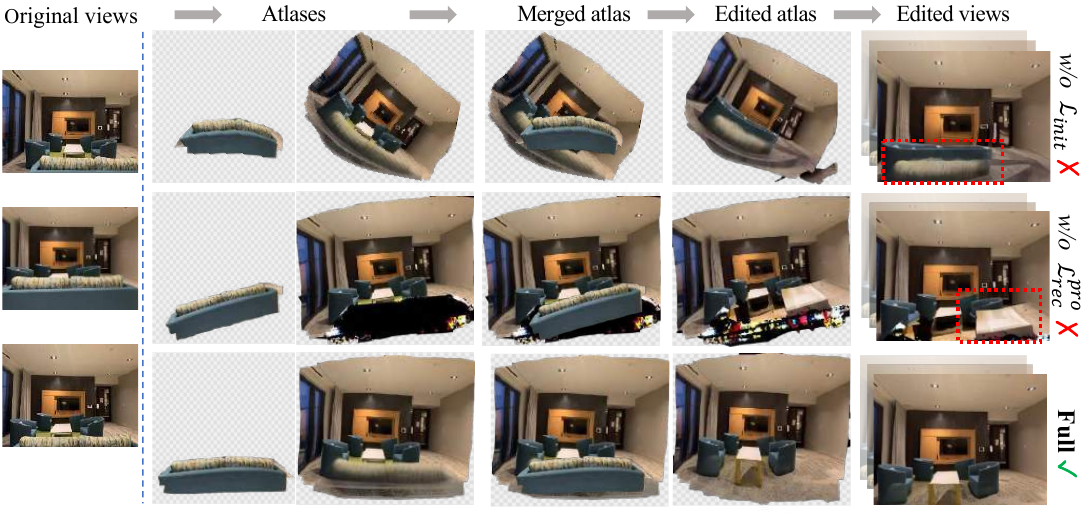}
        \label{fig:ablation_hash}
    }
    \vspace{1mm}
    \hfill
    \subfloat[\textbf{Designed component ablation on the editing strategy and dialogue system.} \textit{SSN} denotes sensitivity to scene names.
    ]{
        \includegraphics[width=0.95\textwidth]{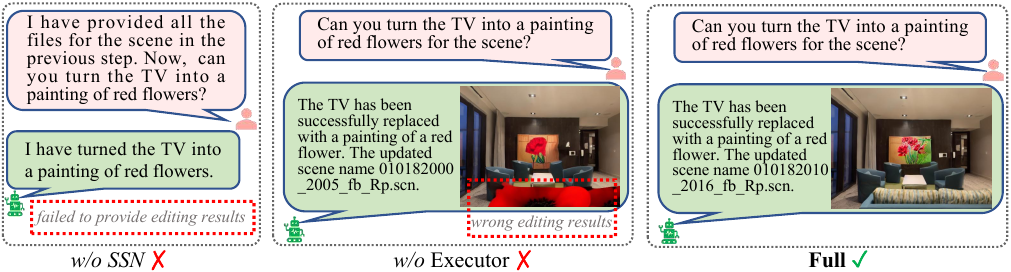}
        \label{fig:ablation_llm}
    }
    \caption{\textbf{Qualitative ablation studies of \modelname{}}. The red box marks failed results.}
    \label{fig:ablation}
\end{figure*}

\subsection{Editing Results and Comparisons}
\noindent{\textbf{Editing Cases of Multiple Rounds Dialogue.}}
In Fig.~\ref{fig:chat-1}, we present a 12-round dialogue example where users provide free-form text queries, and \modelname{} responds with user text, images, or edited results. 
These dialogues involve various editing types, such as object removal or replacement, text-driven or image-driven style transfer, depth map prediction, and scene regeneration based on text and depth map conditions. 
Furthermore, it can also accomplish tasks such as VQA related to a scene and basic text-based dialogues. 
Due to space limitations, the examples provided only partially show the capabilities of \modelname{}, and more results can be found at \url{https://github.com/Fangkang515/CE3D}. 
\revisioncolor{Our approach is broadly compatible with a variety of 2D visual models}, enabling a rich and diverse range of editing effects.
Furthermore, as advancements in 2D visual technologies continue, \modelname{} can dynamically extend its editing capabilities to incorporate newly developed models.

\vspace{1mm}
\noindent{\textbf{Editing Capabilities and Efficiency.}}
\cref{fig:compare_editing} compares the editing capabilities of \modelname{} against state-of-the-art baseline methods, including the 3D scene editing approaches DN2N~\cite{fang2023DN2N} RoMaP~\cite{kim2025RoMaP} and EditSplat~\cite{lee2025editsplat}, as well as the 4D scene editing methods CTRL-D~\cite{CTRL-D}, Instruct-4DGS~\cite{kwon2025Instruct-4DGS} and Dynamic-eDiTor~\cite{lee2025dynamiceditor}.
The results demonstrate the superiority and comprehensiveness of \modelname{}. 
Owing to the workflow decoupled design of 2D editing and 3D/4D reconstruction process, our approach enables the integration of a broader range of editing functionalities, such as sketching human poses, enhancing scene clarity, defocusing the scene, and adjusting colors. 
However, the baseline methods, constrained by their reliance on a single 2D editing model, do not support these diverse editing tasks. 
\revisioncolor{We further conduct quantitative comparisons by applying different editing prompts to each scene across various datasets and calculating the CLIP similarity, CLIP Directional Score, editing time, and peak GPU memory consumption (VRAM peak). Since our framework can flexibly incorporate an arbitrary number of external tools, for a fair comparison, when measuring GPU memory consumption, we load only the external tools required for the corresponding editing task, following the same setting used by the baseline methods.
Table~\ref{tab:compare_clip} shows that our method excels in matching intended text descriptions, due to its enhanced editing versatility. Our method also achieves competitive performance in terms of both editing efficiency and GPU memory consumption.}

\begin{table}[t]
    \centering
    \caption{\textbf{Quantitative ablation studies of \modelname{}}. Here ``$\times$'' indicates that the model variant fails to provide edited results, and ``$-$'' denotes that the evaluation is not applicable. The CLIP-S calculates the similarity between edited results and text. CDS means CLIP Directional Score~\cite{haque2023instruct}, which measures how much the change in text captions agrees with the change in images.}
    \setlength\tabcolsep{3pt}
    \resizebox{0.47\textwidth}{!}{
    \begin{tabular}{c|cccccc}
\toprule
      & w/o $\mathcal{L}_{init}$ & w/o $\mathcal{L}_{rec}^{pro}$ & w/o $\mathcal{L}_{motion}$ & w/o Excutor & w/o SSN & \textbf{Full} \\
\cmidrule{2-7}      & \multicolumn{6}{c}{\textbf{LLFF dataset}} \\
\midrule
CLIP$\uparrow$ & 0.242 & 0.267 & $-$   & 0.190 & $\times$ & \textbf{0.304} \\
CDS$\uparrow$ & 0.163 & 0.179 & $-$   & 0.154 & $\times$ & \textbf{0.192} \\
\midrule
      & \multicolumn{6}{c}{\textbf{CE3D-collect dataset}} \\
\cmidrule{2-7}CLIP$\uparrow$ & 0.286 & 0.263 & $-$   & 0.249 & $\times$ & \textbf{0.352} \\
CDS$\uparrow$ & 0.214 & 0.197 & $-$   & 0.201 & $\times$ & \textbf{0.248} \\
\midrule
      & \multicolumn{6}{c}{\textbf{DynamicNeRF dataset}} \\
\cmidrule{2-7}CLIP$\uparrow$ & 0.270 & 0.256 & 0.295 & 0.237 & $\times$ & \textbf{0.320} \\
CDS$\uparrow$ & 0.193 & 0.184 & 0.198 & 0.188 & $\times$ & \textbf{0.214} \\
\bottomrule
\end{tabular}%

    }
    \label{tab:ablation}
\end{table}

\begin{table}[t]
    \centering
  \setlength{\tabcolsep}{1.5pt}
\caption{\textbf{\revisioncolor{Fine-grained tool-calling failure-rate for Qwen-14B model under different settings.}}}
    \resizebox{0.5\textwidth}{!}{
    \begin{tabular}{c|c|ccc}
\toprule
\textbf{Category} & \textbf{Sub-Category} & \textbf{Original} & \textbf{Few-shot} & \textbf{Trajectory-tuned} \\
\midrule
\multirow{4}[2]{*}{Type} & Understanding & 2\%      & 1\%      & \textbf{0\%} \\
      & Generation & 5\%      & 4\%      & \textbf{1\%} \\
      & Enhancement & 7\%      & 5\%      & \textbf{3\%} \\
      & Conditional Editing & 17\%     & 14\%     & \textbf{7\%} \\
\midrule
\multirow{5}[2]{*}{I/O} & I2T   & 2\%      & 1\%      & \textbf{0\%} \\
      & T2I   & 5\%      & 4\%      & \textbf{1\%} \\
      & I2I   & 6\%      & 6\%      & \textbf{3\%} \\
      & IT2I  & 12\%     & 8\%      & \textbf{5\%} \\
      & RII2I & 6\%      & 5\%      & \textbf{2\%} \\
\midrule
\multirow{3}[2]{*}{Complexity} & Easy  & 3\%      & 1\%      & \textbf{0\%} \\
      & Medium  & 7\%      & 5\%      & \textbf{1\%} \\
      & Hard  & 21\%     & 18\%     & \textbf{10\%} \\
\bottomrule
\end{tabular}%
    }
    \label{tab:rebuttal_llm_rate}
\end{table}

\begin{table}[ht]
    \centering
    \setlength{\tabcolsep}{2pt}
    \caption{\textbf{\revisioncolor{Impact of different factors on 4D atlas quality}.}}
    \resizebox{0.5\textwidth}{!}{
    \begin{tabular}{cc|cc|cc}
\toprule
Atlas resolution & PSNR $\uparrow$ & Video length & PSNR $\uparrow$ & Pivot-view robustness & PSNR $\uparrow$ \\
\midrule
250$\times$250 & 25.17 & 10    & 33.64 & first frame & 28.02 \\
500$\times$500 & 27.59 & 30    & 32.35 & median frame & 27.93 \\
1000$\times$1000 & 28.21 & 60    & 29.82 & last frame & 27.86 \\
1500$\times$1500 & 28.35 & 120   & 27.46 & Pivot frame (ours) & 28.21 \\
\midrule
Moving object count & PSNR $\uparrow$ & Tracking quality & PSNR $\uparrow$ & Complexity & PSNR $\uparrow$ \\
\midrule
1     & 28.85 & Low   & 25.32 & Level 1 & 29.32 \\
2     & 28.02 & Medium & 28.21 & Level 2 & 27.75 \\
3     & 26.46 & High  & 27.94 & Level 3 & 24.77 \\
\bottomrule
\end{tabular}%

    }
    \label{tab:rebuttal_4d_atlas_hyperParams}
\end{table}

\begin{figure*}[t]
    \centering
    \includegraphics[width=1.0\textwidth]{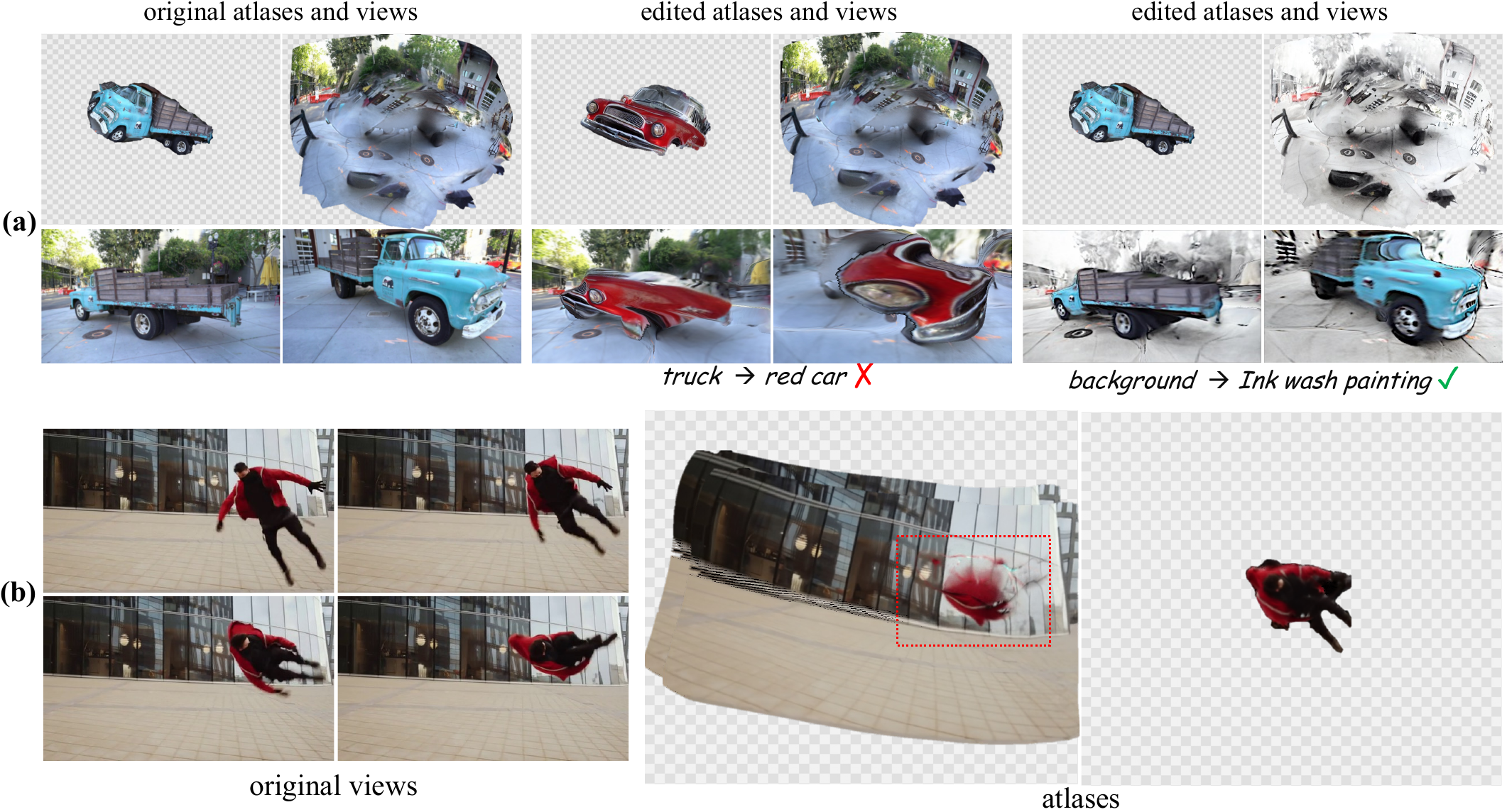}
\caption{\textbf{Failure cases}. (a) In 360-degree scenes, atlas distortions may reduce compatibility with 2D visual models, potentially leading to failed editing results. (b) Under rapid non-rigid deformations or severe occlusions, the tracking model may fail to maintain consistent trajectories, resulting in artifacts in the generated atlas.}
\label{fig:failure_cases}
\end{figure*}

\vspace{1mm}
\noindent{\textbf{Text-query Diversity and Multi-round Dialogue.}}
Furthermore, \cref{fig:compare-IN2N}\textcolor{blue}{(a)} highlights the advantages and robustness of \modelname{} over IN2N~\cite{haque2023instruct} when handling diverse textual queries.
Our method leverages LLMs to interpret user instructions, and its strong language understanding capability helps mitigate the instability of editing results caused by variations in textual input.
\cref{fig:compare-IN2N}\textcolor{blue}{(b)} compares the multi-turn dialogue editing performance of \modelname{} and IN2N across four rounds of interaction.
Due to limited editing functionality and inadequate text interpretation capabilities, the baseline method struggles with complex multi-turn dialogue tasks. 
In contrast, \modelname{} effectively addresses these challenges, demonstrating its strong potential for stable, dialogue-driven interactive editing.
\subsection{Effectiveness of Trajectory-tuning}
We evaluate the performance of several popular LLMs that serve as agents within our proposed \modelname{} framework. 
For a textual instruction, we consider an execution successful only when all external tool invocations are correct; any incorrect tool call is recorded as a failure. 
\cref{fig:compare_llm_rate} shows that mainstream LLMs are generally capable of performing the 3D and 4D editing tasks in our framework, while smaller models such as LLaMA-8B and Qwen-14B exhibit noticeable performance gaps. 
After fine-tuning with our trajectory data, these smaller models improve significantly. Qualitative comparisons are shown in~\cref{fig:compare_qwen}. 
\revisioncolor{We further evaluate the generalization capability of trajectory tuning across different tool characteristics and reasoning complexities. 
The evaluated tools are categorized along three dimensions: \textit{tool type} (Scene Understanding, Generation, Enhancement, and Conditional Editing), \textit{I/O structure} (Image-to-Text: I2T, Text-to-Image: T2I, Image-to-Image: I2I, Image-and-Text-to-Image: IT2I, and Reference-Image-and-Image-to-Image: RII2I), and \textit{reasoning complexity}, with Easy ($\leq2$ turns), Medium (3--5 turns), and Hard ($>5$ turns). 
Table~\ref{tab:rebuttal_llm_rate} reports the corresponding tool-calling failure rates for the original, few-shot, and trajectory-tuned Qwen-14B models.}
Notably, the improvements are most evident for complex, multi-step instructions, where the model must decompose user intent, invoke multiple tools in the correct order, and reuse intermediate results. In such settings, errors at early stages tend to propagate and result in cascading failures.
Furthermore, comparisons with few-shot prompting show no significant gains in tool scheduling accuracy. This is primarily due to the limited context window: the agent prompt already includes extensive tool descriptions and system instructions, and multi-turn interactions further increase the context length. Incorporating few-shot examples for a large set of tools introduces additional overhead, which can lead to attention fragmentation or degraded adherence to constraints in smaller models.
In contrast, trajectory tuning internalizes the planning strategy into the model parameters, enabling more robust and efficient tool-use reasoning without relying on long in-context demonstrations. This makes it particularly well-suited for low-cost and local deployment scenarios.

\subsection{Ablation Studies} \label{subsec:ablation}
We edit scenes from the LLFF, CE3D and DynamicNeRF datasets and quantitatively evaluate the editing results in Table~\ref{tab:ablation}, which demonstrates the importance of each component in achieving dialogue-based high-quality and high-reliability editing.
We qualitatively illustrate the significance of each component in our method in Fig.~\ref{fig:ablation}.
The absence of loss $\mathcal{L}_{init}$ and $\mathcal{L}_{rec}^{pro}$ leads to a decrease in quality and rationality in the edited results. 
The Sensitivity to Scene Names (\textit{SSN}) undertakes the input and output management of scene names and related files.
Removing it and directly using the original file paths as inputs would cause \modelname{} to fail to complete edits correctly and provide users with reasonable responses.
The Executor highlights the importance of managing the merging and separation between foreground and background atlases when editing the atlases. Editing the two atlases independently without the use of the Executor may lead to erroneous editing outcomes due to the lack of complete scene information.

\revisioncolor{We further systematically evaluate the robustness of 4D atlas construction with respect to atlas resolution, video length, dynamic-scene complexity, the number of moving objects, tracking quality, and pivot-view selection. 
The quantitative results are summarized in Table~\ref{tab:rebuttal_4d_atlas_hyperParams}.
Increasing the atlas resolution from $250\times250$ to $1500\times1500$ substantially improves reconstruction quality at lower resolutions, after which the performance gradually saturates. 
We therefore adopt $1000\times1000$ as a practical trade-off between quality and computational cost. 
For video length, the reconstruction quality remains relatively stable for shorter sequences but gradually degrades as the sequence becomes substantially longer, due to the increasing difficulty of establishing reliable and temporally consistent correspondences. 
Similarly, the atlas performs reliably on relatively simple dynamic scenes (Level 1), while severe occlusions (Level 2), large motion, and multiple independently moving objects (Level 3) lead to increasingly challenging correspondence and representation.
We further investigate tracking quality by varying the confidence threshold of CoTracker3~\cite{cotracker3}. 
The results show that inaccurate tracking leads to degradation in atlas quality, highlighting the importance of reliable temporal correspondences. 
Finally, we compare our pivot-view selection strategy with manually selected first, median, and last frames. 
Our heuristic achieves consistently better reconstruction quality, demonstrating that selecting a frame with a larger visible object area provides more informative structural coverage for atlas construction.
Overall, these results show that the proposed 4D atlas performance is naturally affected by long sequences, severe occlusions, rapid motion, and inaccurate tracking.}

\subsection{Discussions and Limitations}

Despite \modelname{}'s capacity for interactive adjustments in 3D and 4D environments, several limitations remain.
These include a reliance on the correctness of LLMs for text query parsing and the need for manual prompt tuning for different visual models.
The effectiveness of scene editing also depends on the performance of external visual tools.
In addition, \modelname{} is primarily optimized for forward-facing and object-centric scenes. In more complex scenarios involving multiple dynamic objects entering or exiting the frame, the pivot-view selection strategy may become ambiguous, potentially leading to unstable atlas representations.
Furthermore, the atlas obtained on 360-degree scenes often exhibits severe distortion, and the training data for existing 2D visual models lack such atlas modality, potentially leading to unreasonable or unsatisfactory editing results (Fig.~\ref{fig:failure_cases}(a)).
Finally, the robustness of the pipeline depends on reliable tracking. In cases of rapid non-rigid deformation or severe occlusion, the tracking model (e.g., CoTracker3) may fail to maintain consistent trajectories, resulting in artifacts in the generated atlas (Fig.~\ref{fig:failure_cases}(b)).

\section{Conclusions} \label{sec:conclusion}
This work addresses the limitations of existing methods that couple 3D or 4D scene representation with 2D image editing models and the challenges they face in enabling more effective interactive designs. 
We demonstrate that representing scenes as 2D atlases (mapping from 3D or 4D scene views to 2D plane images) facilitates compatibility between scene editing and \revisioncolor{a wide range of 2D visual models}. 
By leveraging LLMs to manage visual tools and scene files, we endow \modelname{} with rich editing effects and sustainable conversational editing capabilities, and systematically explore and propose solutions to enhance the stability of LLMs.
Despite the potential for further progress in atlas optimization for complex 360-degree environments, this work advances interactive editing of real-world scenes, expanding beyond restricted visual models to enable a more adaptable editing framework.

\ifCLASSOPTIONcaptionsoff
  \newpage
\fi

{
\bibliographystyle{IEEEtran}
\bibliography{IEEEabrv,ref}
}

\vfill

\end{document}